\pdfoutput=1
\documentclass[journal]{IEEEtranTIE}

\usepackage{cite}
\usepackage{graphicx}
\usepackage{xcolor,soul}
\usepackage{tcolorbox}
\soulregister\cite7
\soulregister\citep7
\soulregister\citet7
\soulregister\ref7
\soulregister\pageref7

\usepackage{amssymb}
\usepackage{bm}
\usepackage{algorithm}
\usepackage{algpseudocode}

\usepackage{multirow}
\usepackage{soul}
\usepackage{svg}
\usepackage{float}
\definecolor{revisionColor}{HTML}{0238A8}
\newcommand{\reviewed}[1]{\textcolor{black}{#1}}
\usepackage{hyperref}
\usepackage{booktabs}
\usepackage{xurl}

\usepackage{cuted}
\usepackage{capt-of}

\hypersetup{
    colorlinks=true,
    linkcolor=black,
    citecolor=black,
    urlcolor=black
}

\usepackage{amsmath}
\usepackage{tabularx}

\title{S$^3$KF: \underline{S}pherical \underline{S}tate-\underline{S}pace Kalman Filtering for Panoramic 3D Multi-Object Tracking}

\author{Zhongyuan~Liu, Shaonan~Yu, Jianping~Li,~\IEEEmembership{Senior Member,~IEEE}, Pengfei~Wan, Xinhang~Xu, Pengfei~Wang, Maggie~Y.~Gao, and Lihua~Xie,~\IEEEmembership{Fellow,~IEEE}%
\thanks{This work was supported by NTUitive Gap Fund (NGF-2025-17006).}%
\thanks{Z. Liu, S. Yu, J. Li, P. Wan, and X. Xu are with the School of Electrical and Electronic Engineering, Nanyang Technological University, Singapore. P. Wang is with ST Engineering, Singapore. M. Gao is with the School of Civil and Environmental Engineering, Nanyang Technological University, Singapore. L. Xie is with NTU–VinUni Joint Research Laboratory for Robotics and Embodied AI, School of Electrical and Electronic Engineering, Nanyang Technological University, Singapore 639798 and VinUniversity, Hanoi 100000, Vietnam. (e-mail: \nolinkurl{zliu051@e.ntu.edu.sg}; 
\nolinkurl{shaonan001@e.ntu.edu.sg}; 
\nolinkurl{jianping.li@ntu.edu.sg}; 
\nolinkurl{pengfei008@e.ntu.edu.sg}; 
\nolinkurl{xu0021ng@e.ntu.edu.sg}; 
\nolinkurl{pengfei.wang@stengg.com}; 
\nolinkurl{yuan008@e.ntu.edu.sg}; 
\nolinkurl{elhxie@ntu.edu,sg; lihua.x@vinuni.edu.vn}). (Z. Liu and S. Yu contributed equally. J. Li is the corresponding author.)}%
}

\begin{document}
% R2: print author names explicitly for consecutive references with identical
% authors (reference [16]) instead of the IEEEtran repeated-author dash.
\bstctlcite{IEEEexample:BSTcontrol}
\maketitle

% Reduce the gap between author block and Fig. 1.

\begin{strip}
\centering
\vspace{-3cm}
\includegraphics[width=\textwidth,height=\textheight,keepaspectratio]{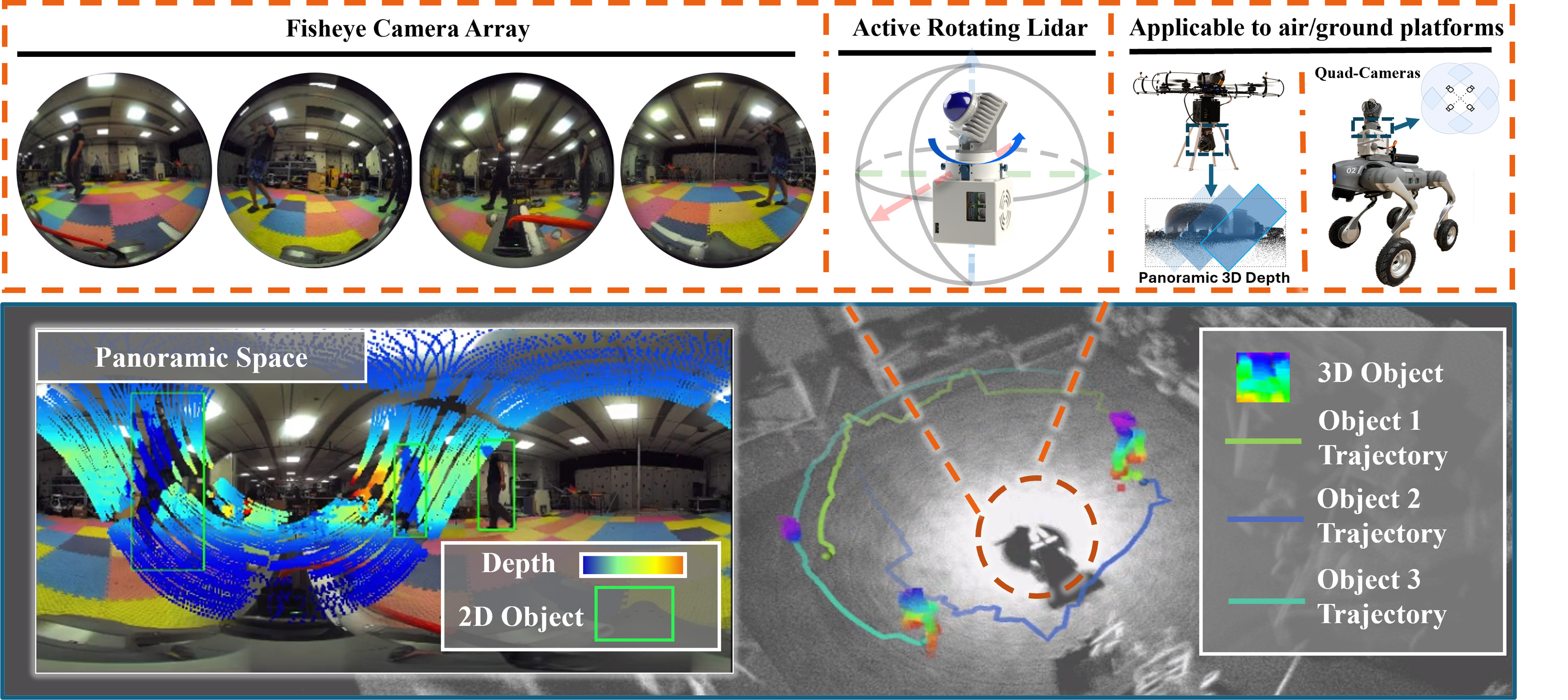}
\captionof{figure}{System overview of the proposed 3D panoramic multi-object tracking framework. A fisheye camera array provides panoramic 2D detections, while an actively rotating LiDAR supplies full \texorpdfstring{360$^{\circ}$}{360 deg} depth coverage. Both modalities are fused in a geometry-consistent state representation on the unit sphere \texorpdfstring{$\mathbb{S}^2$}{S2} using a two-degree-of-freedom tangent-plane parameterization. This formulation eliminates projection singularities of panoramic imagery and unifies 2D and 3D tracking.}
\label{fig:abstract}
\end{strip}

%%%%%%%%%%%%%%%%%%%%%%%%%%%%%%%%%%%%%%%%%%%%%%%%%%%%%%%%%%%%%%%%%%%%%%%%%%%%%%%%

\begin{abstract}

% Panoramic 3D perception is essential for robust multi-object tracking in complex environments. 
% We present a unified framework that integrates a motorized rotating LiDAR with a quad-camera rig, providing synchronized full-coverage geometry and appearance information. 
% Unlike conventional image-based trackers that model states in the 2D image plane and suffer from projection singularities in panoramic imagery, or 3D trackers that rely on redundant Euclidean representations, we propose a geometry-consistent state formulation on the unit sphere $\mathbb{S}^2$. 
% By parameterizing object direction with only two degrees of freedom in the tangent plane, our method avoids over-parameterization and numerical instability, while offering a unified representation that naturally generalizes to both 2D and 3D tracking scenarios. 
% Inspired by ByteTrack, we further augment the state with object width, height, and their velocities, and incorporate depth and depth velocity from LiDAR observations. 
% An extended spherical Kalman filter is derived to jointly estimate direction, scale, and depth, enabling principled fusion of multimodal measurements. 
% Experiments demonstrate that the proposed system achieves accurate and robust panoramic multi-object tracking, effectively handling wide FoV distortions and occlusions while maintaining real-time performance.

Panoramic multi-object tracking is important for industrial safety monitoring, wide-area robotic perception, and deployment in large workspaces. In these settings, the sensing system must provide full-surround coverage, metric geometric cues, and stable target association under wide field-of-view distortion and occlusion. Existing image-plane trackers are tightly coupled to the camera projection and become unreliable in panoramic imagery, while conventional Euclidean 3D formulations introduce redundant directional parameters and do not naturally unify angular, scale, and depth estimation. In this paper, we present $\mathbf{S^3KF}$, a panoramic 3D multi-object tracking framework built on a motorized rotating LiDAR and a quad-fisheye camera rig. The key idea is a wrap-continuous, geometry-consistent state representation on the unit sphere $\mathbb{S}^2$, where object bearing is modeled by a two-degree-of-freedom tangent-plane parameterization and jointly estimated with an effective meridional extent and distance dynamics, so that per-frame LiDAR depth can be fused as a joint observation and the distance estimate is stabilized even when depth is temporarily unavailable. Based on this state, we derive an extended spherical Kalman filtering pipeline that fuses panoramic camera detections with LiDAR depth observations for multimodal tracking. We further establish a map-based ground-truth generation pipeline using wearable localization devices registered to a shared global LiDAR map, enabling quantitative evaluation without motion-capture infrastructure. Experiments on self-collected real-world sequences show decimeter-level 3D tracking accuracy, zero identity switches over all five sequences, consistent improvements over 2D image-plane and 3D Cartesian-state baselines under intermittent depth observations, and real-time onboard operation on a Jetson AGX Orin platform. These results indicate that the proposed framework is a practical solution for panoramic multi-object tracking with potential for industrial safety monitoring. The project page can be found at \url{https://kafeiyin00.github.io/S3KF/}.

\end{abstract}


%%%%%%%%%%%%%%%%%%%%%%%%%%%%%%%%%%%%%%%%%%%%%%%%%%%%%%%%%%%%%%%%%%%%%%%%%%%%%%%%
\section{Introduction}

% Panoramic 3D perception is a key enabler for robust multi-object tracking (MOT) in complex environments, spanning mobile robotics \cite{jin2024gs} and autonomous driving \cite{liao2024mobile,Singh_2023_ICCV,wen2023panacea} through to wide-area surveillance \cite{li2025graph}. 
% In practice, it is crucial to maintain accurate trajectories of dynamic targets across large fields of view (FoV), under occlusion and across rapid viewpoint changes\cite{li2026aeos}. 
% Camera-only systems provide high-resolution appearance cues but lack reliable metric depth\cite{fischer2022ccdt}; fixed-FoV LiDARs offer precise geometry yet leave blind spots that degrade tracking continuity in full-surround scenes. 
% By contrast, a motorized (rotating) LiDAR mounted on a rotary stage can deliver 360° depth coverage \cite{li2026aeos}, while a multi-camera rig supplies dense appearance and robust detections; integrating these complementary modalities opens up new possibilities for panoramic MOT~\cite{geiger2012cvpr,caesar2020cvpr,yan2023spinloam,li2025ua}.

Panoramic 3D perception is increasingly required in industrial monitoring, field robotics, and wide-area autonomous systems, where targets must be tracked over the full surrounding scene rather than within a narrow forward-facing field of view (FoV) \cite{jin2024gs,liao2024mobile,Singh_2023_ICCV,wen2023panacea,li2025graph}. In such environments, reliable tracking of workers and moving equipment is directly related to collision prevention and safe human--robot coexistence \cite{zhou2021intelligent,wan2019blockchain}, and a practical system must provide full-surround sensing, metric geometric awareness, and embedded real-time operation under clutter, occlusion, and frequent viewpoint changes.

Meeting these requirements with a single sensing modality remains difficult: camera-only systems provide rich appearance cues but lack reliable metric depth, which makes association fragile when objects overlap or undergo large scale changes \cite{fischer2022ccdt,yang2026agi2p}; fixed-FoV LiDARs offer accurate geometry but leave blind spots that interrupt full-surround tracking. A motorized rotating LiDAR can recover panoramic depth coverage from a single unit \cite{li2026aeos}, while a multi-camera rig supplies dense visual observations; however, simply combining these sensors does not resolve the representation problem at the core of panoramic tracking.

Most tracking-by-detection pipelines model object states directly in the image plane, which becomes problematic for fisheye or equirectangular views, where projection distortion is highly nonuniform and singular behavior appears near the poles \cite{zhang2022bytetrack,sun2024smmpod,Ong}. Euclidean 3D formulations instead represent target direction with three parameters plus a unit-norm constraint, which is redundant for bearing estimation and complicates filtering and fusion \cite{weng2020ab3dmot}. Moreover, many practical multi-object tracking (MOT) systems treat depth and physical scale as auxiliary cues rather than first-class state variables, limiting robustness when appearance is ambiguous or temporarily missing.

This paper addresses the above limitations with a unified panoramic tracking framework, $\mathbf{S^3KF}$, that couples a rotating LiDAR and a quad-fisheye camera rig with a geometry-consistent state space on the unit sphere $\mathbb{S}^2$. Instead of tracking in the distorted image plane, we represent object bearing by a two-degree-of-freedom tangent-plane parameterization and jointly model directional motion, object scale, and depth dynamics. This representation is wrap-continuous across the panorama seam, avoids image-space singularities, removes redundant directional coordinates, and provides a common interface for panoramic 2D and depth-augmented 3D tracking, on which we build an extended spherical Kalman filtering pipeline that fuses camera detections and LiDAR depth observations while preserving the underlying geometry of the bearing state.

Beyond the filtering formulation, we address a practical bottleneck in evaluating panoramic 3D human tracking: accurate ground truth is difficult to obtain in large real environments without motion-capture infrastructure. We therefore establish a map-based trajectory acquisition pipeline in which wearable localization devices register online LiDAR scans to a shared global map, producing globally consistent reference trajectories for multiple targets.

Our main contributions are fourfold:
\begin{itemize}
    \item We develop a panoramic sensing platform that integrates a motorized rotating LiDAR with a quad-fisheye camera rig, providing synchronized full-surround geometric and appearance observations for embedded multi-object tracking.
    \item We propose a geometry-consistent spherical state representation for panoramic tracking, where target bearing is parameterized on $\mathbb{S}^2$ using two tangent-plane coordinates that remain continuous across the panorama wrap-around seam and is jointly modeled with an effective meridional extent and distance dynamics.
    \item We derive an extended spherical Kalman filtering framework that fuses panoramic image detections and per-frame LiDAR depth in a unified state-space model: the LiDAR depth enters the filter as a joint observation with an explicitly modeled scale--depth cross-covariance, the angular height couples the effective meridional extent with the distance, providing a geometric constraint on the latter during image-only segments, and a blind-zone gain-decay mechanism regularizes the scale direction during long depth outages. A controlled ablation attributes the accuracy and identity gains to these fusion mechanisms and to the wrap-continuity of the angular state.
    \item We establish a map-based ground-truth generation pipeline with wearable localization devices in a shared global LiDAR map, enabling infrastructure-light quantitative evaluation of panoramic 3D human tracking in large indoor and outdoor environments.
\end{itemize}

By jointly modeling direction, scale, and depth, our framework stabilizes association across large FoV changes and occlusions, while depth aggregation from the rotating LiDAR provides accurate metric information.
The code, calibration files, evaluation scripts, and self-collected sequences with ground-truth trajectories will be publicly released on the project page upon publication.

\section{Related Works}

\subsection{Hardware Systems for Panoramic Perception}
Panoramic perception has long been pursued via omnidirectional imaging and multi-camera rigs, leveraging central panoramic models on the sphere~\cite{geyer2000eccv,geyer2001ijcv} and practical calibration toolchains~\cite{scaramuzza2006iros,kiode}. Camera-only systems provide wide FoV and rich appearance cues, but depth remains ambiguous, whereas modern LiDAR systems deliver metrically accurate 3D geometry, with 360$^{\circ}$ sensor suites promoted by datasets such as KITTI and nuScenes~\cite{geiger2012cvpr,caesar2020cvpr}. To reduce blind spots without duplicating sensors, motorized (rotating) LiDARs mount a multi-beam scanner on a rotary stage, achieving panoramic coverage from a single unit; recent works demonstrate SLAM/odometry feasibility with such actuated configurations~\cite{li2025limo,li2025ua}, and multi-LiDAR platforms achieve similar coverage via sensor multiplicity and online extrinsic calibration~\cite{jiao2021mloam}. However, integrating a motorized LiDAR with a multi-camera rig \emph{specifically for robust panoramic multi-object tracking} remains underexplored. Our platform combines a rotating LiDAR with a quad-camera rig to deliver synchronized, full-coverage depth and appearance.

\subsection{Geometric State Representation for Tracking}
Conventional multi-object trackers parameterize targets either in the 2D image plane, e.g., ByteTrack with bounding-box centers and velocities in pixel coordinates~\cite{zhang2022bytetrack} or CenterTrack with keypoints and their frame-to-frame displacement \cite{zhou2020}, or in 3D Euclidean space, where directions require explicit unit-norm constraints and redundant degrees of freedom complicate filtering~\cite{weng2020ab3dmot}. Image-space states are tied to the projection model and cannot be generalized to panoramic imagery, where equirectangular representations suffer from severe distortions and singularities at the poles~\cite{sun2024smmpod}. In this work, we adopt a geometry-consistent state representation directly on the unit sphere $\mathbb{S}^2$, where object directions are expressed with a two-degree-of-freedom tangent-plane parameterization. This formulation eliminates the over-parameterization of Euclidean models, avoids image-plane singularities, and provides a unified interface between 2D and 3D domains within a single state-space framework.

\subsection{Ground Truth Generation for 3D Tracking}

Reliable ground-truth trajectories are essential for quantitative evaluation of 3D tracking systems. Indoor benchmarks typically rely on optical motion-capture systems (e.g., TUM RGB-D~\cite{sturm2012iros}, EuRoC MAV~\cite{burri2016ijrr}), which provide millimeter-level accuracy but require extensive infrastructure and are limited to controlled indoor environments. Outdoor datasets commonly use RTK-GNSS receivers~\cite{li2023whu,li2024hcto}, as in KITTI~\cite{geiger2012cvpr} and nuScenes~\cite{caesar2020cvpr}, but their accuracy degrades under occlusion and multipath, and wearable deployment on multiple human subjects is often impractical.

An alternative paradigm is map-based localization \cite{li2025graph}. Building on this approach, we develop a map-matching-based framework for ground-truth generation. Each wearable localization device independently registers its online LiDAR observations to a pre-built shared global map, yielding globally consistent 6-DoF trajectories without requiring external infrastructure or inter-device calibration. This setup provides an accurate, repeatable, and scalable solution for ground-truth generation in outdoor multi-person 3D tracking under realistic deployment conditions.

\section{\texorpdfstring{$\mathbf{S^3KF}$}{S3KF}: Spherical State-Space Kalman Filtering}
\begin{figure}[]
\centering
\includegraphics[width=1.0\linewidth]{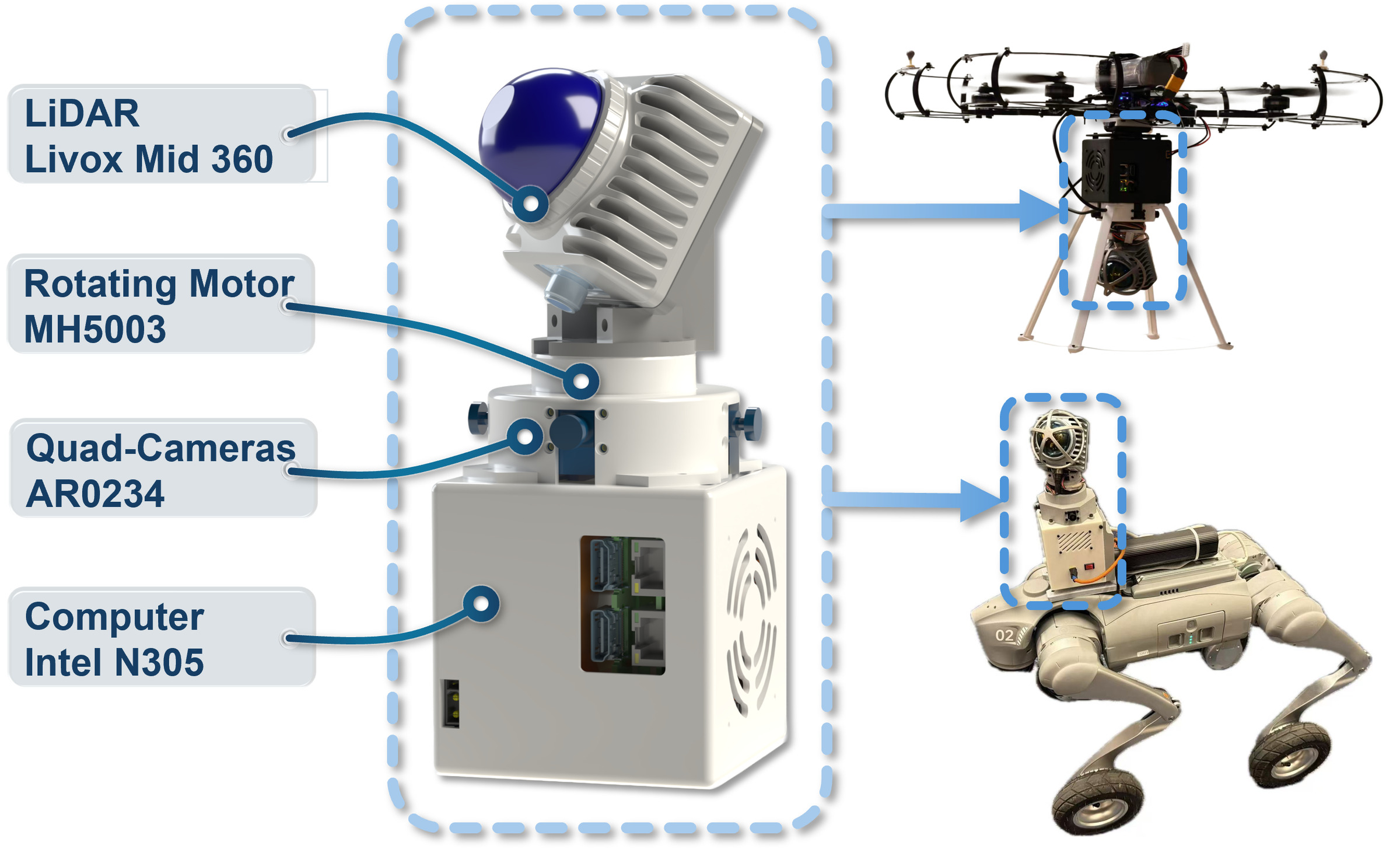}
    \caption{Hardware of the proposed panoramic sensing system.}
    \label{fig:hardware}
\end{figure}

\begin{figure}[]
\centering
\includegraphics[width=\linewidth]{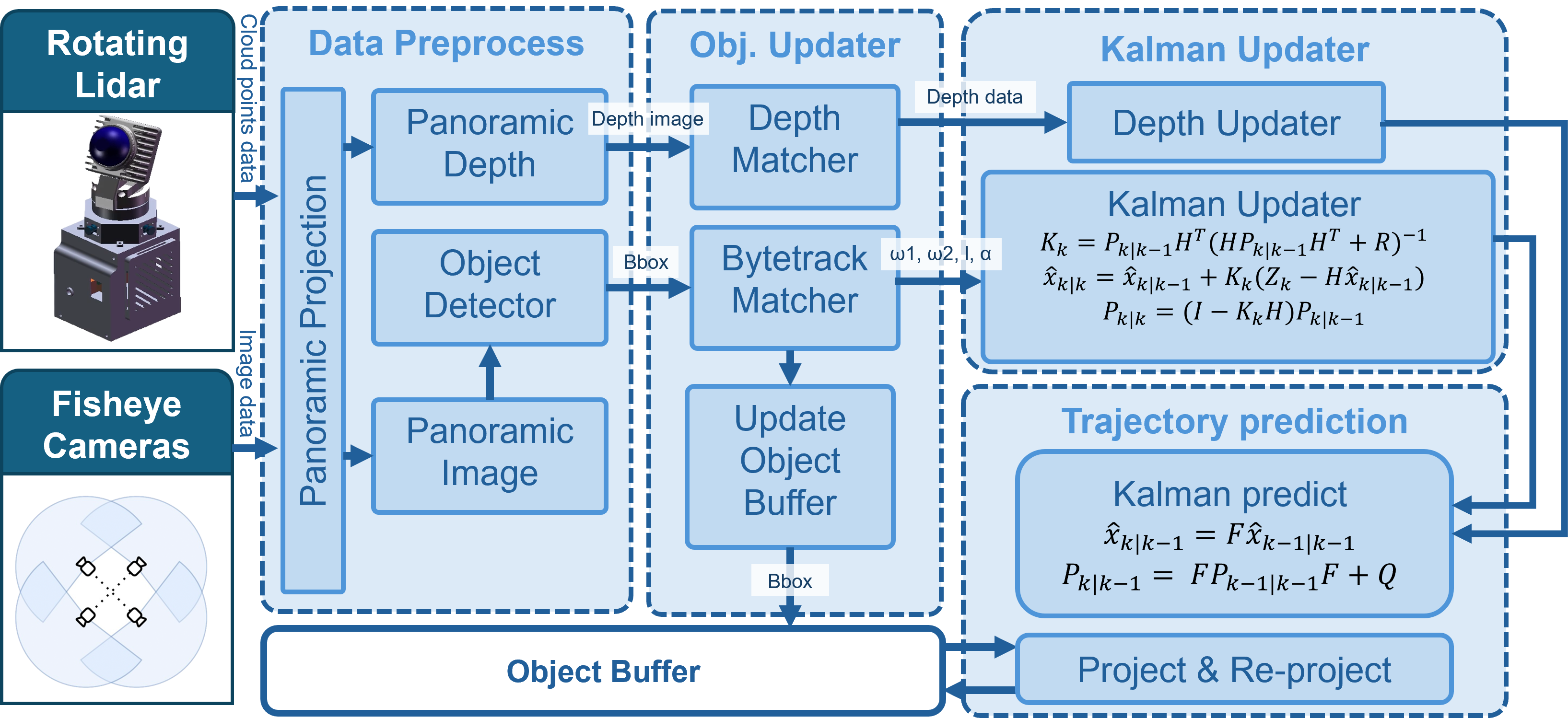}
    \caption{Workflow of the proposed panoramic 3D multi-object tracking. Arrows indicate the data flow: the quad-fisheye camera array provides panoramic 2D detections (target bearing and bounding box), and the rotating LiDAR provides metric depth measurements; both modalities are transformed into the unified spherical state representation and fused in the extended spherical Kalman filter for joint state estimation and data association.}
    \label{fig:Workflow}
\end{figure}

\subsection{Problem Statement and Notation}
\begin{figure}[]
\centering
\includegraphics[width=0.5\linewidth]{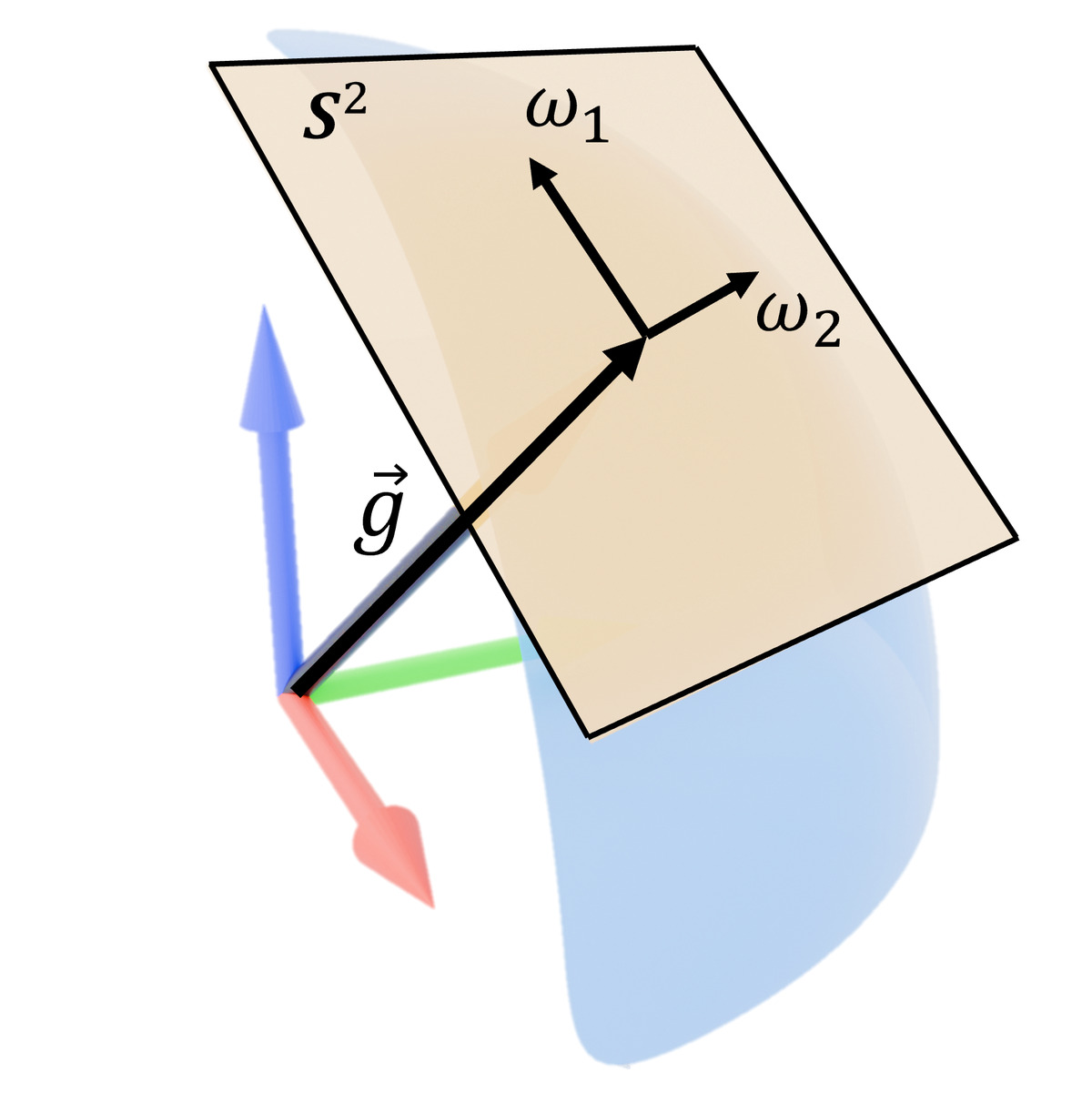}
    \caption{\texorpdfstring{$\mathbf{g} \in \mathbb{S}^2$}{g in S2} denotes the true bearing vector of the object. Because \texorpdfstring{$\mathbb{S}^2$}{S2} is a nonlinear manifold, we employ a local tangent parameterization for filtering.}
    \label{fig:tangent_space}
\end{figure}

\reviewed{The notation is summarized in Table~VII of Appendix~B, which is provided as supplementary material.} The hardware of the proposed panoramic 3D sensing system (Fig.~\ref{fig:hardware}) couples a rotating LiDAR with a quad-fisheye camera rig, and the workflow is shown in Fig.~\ref{fig:Workflow}. We address multi-object tracking in panoramic imagery augmented with depth sensing. Each object direction is represented as a point on the unit sphere
\begin{equation}
\mathbb{S}^2 = \{\mathbf{g}\in\mathbb{R}^3 \,|\, \|\mathbf{g}\|=1\},
\end{equation}
where $\mathbf{g}_k \in \mathbb{S}^2$ denotes the true bearing vector at time step $k$. Because $\mathbb{S}^2$ is a nonlinear manifold, we employ a local tangent parameterization for filtering (Fig.~\ref{fig:tangent_space}): $\bar{\mathbf{g}}_k \in \mathbb{S}^2$ is the current reference estimate of the direction, and the tangent plane at this point is
\begin{equation}
T_{\bar{\mathbf{g}}_k}\mathbb{S}^2 = \{\mathbf{v}\in\mathbb{R}^3 \,|\, \bar{\mathbf{g}}_k^\top \mathbf{v} = 0\}\reviewed{.}
\end{equation}
\reviewed{Here, $\mathbf{B}_k=[\mathbf{b}_{1,k},\mathbf{b}_{2,k}]\in\mathbb{R}^{3\times2}$ is the tangent-basis matrix, whose columns form an orthonormal basis of $T_{\bar{\mathbf{g}}_k}\mathbb{S}^2$ ($\mathbf{B}_k^\top\mathbf{B}_k=\mathbf{I}_2$, $\mathbf{B}_k^\top\bar{\mathbf{g}}_k=\mathbf{0}$) and maps local coordinates $\mathbf{w}_k=[w_{1,k},w_{2,k}]^\top\in\mathbb{R}^2$ to tangent vectors $\mathbf{B}_k\mathbf{w}_k$; conversely, $\mathbf{w}_k=\mathbf{B}_k^\top\mathrm{Log}_{\bar{\mathbf{g}}_k}(\mathbf{g}_k)$;} the explicit log/exp mapping formulas are given in Appendix~\ref{app:logexp}.

\subsection{State Representation}
Traditional trackers such as ByteTrack represent objects with image-plane coordinates and bounding-box size. 
However, panoramic cameras introduce nonlinear distortions, and depth sensing provides complementary 3D information~\cite{geyer2000eccv}. 
We therefore propose an extended spherical state representation:

\begin{equation}
\reviewed{\mathbf{x}_k =
\bigl[w_{1,k},w_{2,k},\dot{w}_{1,k},\dot{w}_{2,k},
\ell_k,\dot{\ell}_k,h_k,\dot{h}_k,d_k,\dot{d}_k\bigr]^\top
\in\mathbb{R}^{10}.}
\label{eq:states}
\end{equation}

\reviewed{In Eq.~\eqref{eq:states}, $(w_{1,k},w_{2,k})$ are local bearing coordinates and $(\dot w_{1,k},\dot w_{2,k})$ their velocities. The dimensionless angular aspect ratio $\ell_k$ is the angular width divided by the angular height; $h_k$ is the effective meridional extent in meters (Section~III-D), and $d_k$ is the range along the bearing in meters. Dotted symbols denote the corresponding velocities. After each prediction and update, the reference bearing is advanced and the tangent coordinates reset to zero, keeping the filter near the local origin and avoiding a panorama-seam discontinuity. Camera detections measure bearing and aspect ratio; angular height and LiDAR depth jointly constrain effective extent and distance.}

\subsection{Motion Model}
We assume constant-velocity dynamics for all components:
\begin{equation}
\mathbf{x}_{k+1} = \mathbf{F}\mathbf{x}_k + \mathbf{q}_k,\quad 
\mathbf{q}_k \sim \mathcal{N}(\mathbf{0},\mathbf{Q}),
\end{equation}
with block-diagonal transition
\begin{equation}
\mathbf{F} = 
\text{diag}(\mathbf{F}_{\text{pos}}, \mathbf{F}_{\ell}, \mathbf{F}_{h}, \mathbf{F}_{d}),
\end{equation}
where $\mathbf{F}_{\text{pos}}$ acts on the position--velocity pairs of $(w_1,w_2)$ in the state ordering of \eqref{eq:states},
\begin{equation}
\mathbf{F}_{\text{pos}}=
\begin{bmatrix}
1 & 0 & \Delta t & 0\\
0 & 1 & 0 & \Delta t\\
0 & 0 & 1 & 0\\
0 & 0 & 0 & 1
\end{bmatrix},
\end{equation}
and $\mathbf{F}_{\ell},\mathbf{F}_{h},\mathbf{F}_{d}$ are $2\times 2$ constant-velocity blocks $[1,\ \Delta t;\ 0,\ 1]$ for $\ell$, $h$, and $d$, respectively, where $\Delta t$ is the sampling interval.
The prediction step is
\begin{align}
\hat{\mathbf{x}}_{k|k-1}&=\mathbf{F}\hat{\mathbf{x}}_{k-1|k-1},\\
\mathbf{P}_{k|k-1}&=\mathbf{F}\mathbf{P}_{k-1|k-1}\mathbf{F}^\top+\mathbf{Q}.
\end{align}

\subsection{Measurement Model}
We fuse panoramic camera detections and depth observations.  

\subsubsection{Direction measurement} 
A detection $\tilde{\mathbf{g}}_k \in \mathbb{S}^2$ is projected to the tangent plane at $\bar{\mathbf{g}}_k$: 
\begin{equation}
\mathbf{p}_k=(\mathbf{I}-\bar{\mathbf{g}}_k\bar{\mathbf{g}}_k^\top)\tilde{\mathbf{g}}_k,\quad
\theta_k=\arccos(\bar{\mathbf{g}}_k^\top \tilde{\mathbf{g}}_k).
\end{equation}
With $\hat{\mathbf{p}}_k=\mathbf{p}_k/\|\mathbf{p}_k\|$, the log-map is
\begin{equation}
\reviewed{\boldsymbol{\xi}_k=\theta_k\hat{\mathbf{p}}_k\in T_{\bar{\mathbf{g}}_k}\mathbb{S}^2\subset\mathbb{R}^3.}
\label{eq:log_vector}
\end{equation}
\reviewed{The three-component column vector $\boldsymbol{\xi}_k$ has local coordinates $[z_{1,k},z_{2,k}]^\top=\mathbf{B}_k^\top\boldsymbol{\xi}_k\in\mathbb{R}^2$.}

\subsubsection{Bounding box and depth}
Panoramic image detections provide the detected bounding box, from which the angular aspect ratio $\ell_k^{\text{det}}$ (dimensionless) and the pixel height $h_k^{\text{det}}$ are derived, while the rotating LiDAR provides a depth observation $d_k^{\text{obs}}$ (in meters) for the matched target.

Under the equirectangular panoramic projection, the vertical pixel coordinate maps linearly to the elevation angle, so the detected pixel height is converted into the \emph{angular height}
\begin{equation}
\alpha_k^{\mathrm{det}}=\frac{h_k^{\mathrm{det}}}{h_{\mathrm{img\_height}}}\,2\varphi_{\mathrm{fov}},
\label{eq:alpha_det}
\end{equation}
where $h_{\mathrm{img\_height}}$ is the panorama height in pixels ($640$ in our implementation) and $2\varphi_{\mathrm{fov}}$ is the total vertical field of view ($\varphi_{\mathrm{fov}}=\pi/2$ for the $360^{\circ}\times 180^{\circ}$ image). Likewise, the detected angular width is $(s_x/W)\,2\pi$ ($s_x$: pixel box width, $W$: panorama width), giving the detected angular aspect ratio
\begin{equation}
\ell_k^{\mathrm{det}}=\frac{s_x}{W}\,\frac{2\pi}{\alpha_k^{\mathrm{det}}}.
\label{eq:ell_det}
\end{equation}
These conversions involve only the current detection and the fixed panoramic geometry, independent of the filter state.

The per-detection LiDAR depth is extracted on the synchronized frame: the point cloud (timestamp within $60$\,ms of the image) is splatted onto the panoramic depth map, and within the core region of each detection box the depth is the median of the \emph{nearest dense cluster} (a $0.3$\,m window with at least $125$ splat points and $15\%$ valid pixels)---the nearest cluster is the body surface, which occludes the background; frames failing this test carry no depth observation.

The angular height is related to the \emph{effective meridional extent} $h_k$ (m) and the distance $d_k$ (m) through
\begin{equation}
\alpha_k=2\arctan\!\Big(\frac{h_k}{2\,d_k}\Big).
\label{eq:alpha_geom}
\end{equation}
Here, $h_k$ represents the target's effective extent along the local meridional direction on the tangent plane, rather than its physical body height.

\paragraph*{Image-only observation}
When only an image detection is available, the image measurement vector is
\begin{equation}
\mathbf{z}_{k,\mathrm{img}}=
\begin{bmatrix}
z_{1,k} & z_{2,k} & \ell_k^{\text{det}} & \alpha_k^{\text{det}}
\end{bmatrix}^{\top},
\label{eq:z_img}
\end{equation}
where $z_{1,k}$ and $z_{2,k}$ are the tangent-plane components of the directional error from the predicted bearing to the detected bearing, and $\ell_k^{\mathrm{det}}$ is the detected angular aspect ratio defined in~\eqref{eq:ell_det}. The measurement update is carried out in two sequential steps. The first three components correspond to $(w_{1,k}, w_{2,k}, \ell_k)$. The scalar angular-height measurement is then incorporated through the nonlinear geometric relation in~\eqref{eq:alpha_geom}. Evaluated at the prior estimate, its predicted value is
\begin{equation}
\hat{z}_{\alpha,k}=2\arctan\!\Big(\frac{\hat{h}_{k|k-1}}{2\,\hat{d}_{k|k-1}}\Big),
\label{eq:z_alpha_pred}
\end{equation}
and the corresponding Jacobian with respect to the \emph{effective meridional extent} $h$ and distance $d$ is
\begin{equation}
\frac{\partial\alpha}{\partial h}=\frac{d}{\Delta},\qquad
\frac{\partial\alpha}{\partial d}=-\frac{h}{\Delta},\qquad
\Delta=d^{2}+\frac{h^{2}}{4}.
\label{eq:jac_alpha}
\end{equation}
Stacking the three linear selections and the Jacobian row of the angular-height measurement yields the effective image observation matrix $\mathbf{H}_{\mathrm{img}}$ of the measurement function $\mathbf{h}_{\mathrm{img}}(\mathbf{x})=[w_{1},w_{2},\ell,\,2\arctan(h/(2d))]^{\top}$, so the corresponding innovation $\mathbf{z}_{k,\mathrm{img}}-\mathbf{h}_{\mathrm{img}}(\hat{\mathbf{x}}_{k|k-1})$ has the nonlinear fourth component $\alpha_k^{\mathrm{det}}-2\arctan(\hat{h}_{k|k-1}/(2\hat{d}_{k|k-1}))$:
\begin{equation}
\mathbf{H}_{\mathrm{img}}=
\begin{bmatrix}
1&0&0&0&0&0&0&0&0&0\\
0&1&0&0&0&0&0&0&0&0\\
0&0&0&0&1&0&0&0&0&0\\
0&0&0&0&0&0&w\,\dfrac{d}{\Delta}&0&-\dfrac{h}{\Delta}&0
\end{bmatrix}.
\label{eq:H_img}
\end{equation}
The first three rows are exact linear selections, while the fourth row is the angular-height Jacobian evaluated at the prior estimate. The corresponding measurement covariance is block diagonal,
$\mathbf{R}_{\mathrm{img}}=\mathrm{diag}(\sigma_{\mathrm{dir}}^{2},\sigma_{\mathrm{dir}}^{2},R_{\ell},R_{\alpha})$.
The weight $w$ is the blind-zone gain-decay factor applied to the height channel (Section~III-E), with $w=1$ when depth has been available recently. It scales both the corresponding Jacobian entry and Kalman gain rows. In implementation, the two blocks are processed sequentially. The linear component is updated without gating, followed by a scalar angular-height update gated on the normalized innovation squared (NIS). When $w=1$, this sequential formulation is equivalent to a single update using~\eqref{eq:H_img} and $\mathbf{R}_{\mathrm{img}}$. Since $\alpha_k^{\mathrm{det}}$ is computed solely from the current detection, while the prior estimate is used only to predict the measurement, the measurement remains conditionally independent of the prior state, consistent with the standard EKF formulation for nonlinear measurements.

\paragraph*{Joint image--LiDAR observation}
When a synchronized LiDAR depth observation is available, the two modalities are fused in a single joint update with
\begin{equation}
\mathbf{z}_k=
\begin{bmatrix}
z_{1,k} & z_{2,k} & \ell_k^{\text{det}} & h_k^{\mathrm{meas}} & d_k^{\mathrm{obs}}
\end{bmatrix}^{\top}\in\mathbb{R}^{5},
\label{eq:z_joint}
\end{equation}
where the effective-scale observation is synthesized from the two \emph{current} measurements (the image angular height and the LiDAR depth) as
\begin{equation}
h_k^{\mathrm{meas}}=2\,d_k^{\mathrm{obs}}\,\tan\!\big(\alpha_k^{\mathrm{det}}/2\big),
\label{eq:h_meas}
\end{equation}
so that no prior state estimate is involved in the measurement construction. All five components are linear selections of the states $(w_{1,k},w_{2,k},\ell_k,h_k,d_k)$, i.e.,
\begin{equation}
\mathbf{z}_k = \mathbf{H}\mathbf{x}_k + \mathbf{r}_k,\quad
\mathbf{r}_k \sim \mathcal{N}(\mathbf{0},\mathbf{R}),
\label{eq:z_Hx}
\end{equation}
with
\begin{equation}
\mathbf{H}=
\begin{bmatrix}
1&0&0&0&0&0&0&0&0&0\\
0&1&0&0&0&0&0&0&0&0\\
0&0&0&0&1&0&0&0&0&0\\
0&0&0&0&0&0&1&0&0&0\\
0&0&0&0&0&0&0&0&1&0
\end{bmatrix}.
\label{eq:H_joint}
\end{equation}
Because $h_k^{\mathrm{meas}}$ and $d_k^{\mathrm{obs}}$ share the same LiDAR depth, their measurement errors are correlated; omitting this correlation would count the depth information twice. The scale--depth block of $\mathbf{R}$ is therefore written in full as
\begin{equation}
\mathbf{R}_{h,d}=
\begin{bmatrix}
\big(\tfrac{\partial h}{\partial d}\big)^{2}R_{d}+\big(\tfrac{\partial h}{\partial\alpha}\big)^{2}R_{\alpha} & \tfrac{\partial h}{\partial d}\,R_{d}\\[2pt]
\tfrac{\partial h}{\partial d}\,R_{d} & R_{d}
\end{bmatrix},
\label{eq:cov_hd}
\end{equation}
with $\partial h/\partial d=2\tan(\alpha_k^{\mathrm{det}}/2)$ and $\partial h/\partial\alpha=d_k^{\mathrm{obs}}/\cos^{2}(\alpha_k^{\mathrm{det}}/2)$, and the joint update is applied as a single 5-dimensional correction rather than as sequential scalar updates, since splitting would discard this correlation.

A nominal scale prior of $1.73$\,m with $25\%$ relative uncertainty is used only for track initialization when no LiDAR depth is available in the first frame. In this case, the initial distance is inferred from the observed angular height as
$d_0=1.73/\big(2\tan(\alpha_k^{\mathrm{det}}/2)\big)$.
Once a depth observation becomes available, the \emph{effective meridional extent} is estimated as a free state from the measurements, and the nominal prior is no longer enforced.

\subsection{Kalman Filter Update and Algorithm}
Given the constant-velocity prediction $(\hat{\mathbf{x}}_{k|k-1},\mathbf{P}_{k|k-1})$ and an available measurement configuration $m$ (the image-only observation \eqref{eq:z_img} or the joint observation \eqref{eq:z_joint}), with measurement function $\mathbf{h}_{m}(\cdot)$ and Jacobian $\mathbf{H}_{m}=\partial\mathbf{h}_{m}/\partial\mathbf{x}$ evaluated at the prior estimate, the correction follows the standard Kalman update on the locally Euclidean state:
\begin{align}
\mathbf{S}_{k,m} &= \mathbf{H}_{m}\mathbf{P}_{k|k-1}\mathbf{H}_{m}^\top + \mathbf{R}_{m}, \\
\mathbf{K}_{k,m} &= \mathbf{P}_{k|k-1}\mathbf{H}_{m}^\top \mathbf{S}_{k,m}^{-1}, \\
\hat{\mathbf{x}}_{k|k} &= \hat{\mathbf{x}}_{k|k-1}+\mathbf{K}_{k,m}\!\big(\mathbf{z}_{k,m}-\mathbf{h}_{m}(\hat{\mathbf{x}}_{k|k-1})\big), \\
\mathbf{P}_{k|k} &= (\mathbf{I}-\mathbf{K}_{k,m}\mathbf{H}_{m})\mathbf{P}_{k|k-1}.
\end{align}
In implementation, a LiDAR depth observation is accepted only if a LiDAR scan is available within a synchronization tolerance of the image frame. When a valid depth observation is available, the update is performed once using the joint observation $(\mathbf{z}_{k},\mathbf{H},\mathbf{R})$ defined in \eqref{eq:z_joint}--\eqref{eq:cov_hd}. If no valid depth observation is available or the joint update is rejected by the gate, the image-only update \eqref{eq:z_img}--\eqref{eq:jac_alpha} is applied instead, ensuring that a depth outlier does not invalidate the image information. All covariance updates use the Joseph form,
$\mathbf{P}_{k|k}=(\mathbf{I}-\mathbf{K}\mathbf{H})\mathbf{P}_{k|k-1}(\mathbf{I}-\mathbf{K}\mathbf{H})^{\top}+\mathbf{K}\mathbf{R}\mathbf{K}^{\top}$,
which remains valid when the gain is scaled by the blind-zone weight defined below.
After the update, we reconstruct the direction on the sphere by the exponential map with the updated local
coordinates $(\hat{w}_{1,k},\hat{w}_{2,k})$, renormalize to unit length, and recompute the tangent basis
$\mathbf{B}_k$ at $\bar{\mathbf{g}}_k$ so that the tangent coordinates are reset to zero and the next step again operates near the origin (Appendix~A). The velocity components in the tangent space are
parallel-transported from the temporary basis via $\mathbf{T}_k=\mathbf{B}_k^\top \mathbf{B}^{\text{tmp}}_k$
to avoid spurious rotation of the local frame.

\subsubsection{Blind-zone gain decay}
Without LiDAR depth, the angular-height measurement constrains only the ratio between $h_k$ and $d_k$. Consequently, the pair $(h_k,d_k)$ is unobservable along $\alpha=\mathrm{const}$, and an image-only update may introduce coupled changes in both states. Since the rotating LiDAR may leave a given bearing without depth observations for up to one rotation period ($\approx 10$\,s), we gradually reduce the influence of the angular-height measurement as the depth-outage duration $t_{\mathrm{blind}}$ increases. Specifically, the weight $w=\exp(-t_{\mathrm{blind}}/\tau)$ is applied to the Jacobian entry in~\eqref{eq:jac_alpha} and the corresponding rows of the Kalman gain. The height velocity is frozen when $t_{\mathrm{blind}}>\tau$, with $\tau=1.0$\,s. As shown by the ablation study in Section~\ref{sec:ablation}, this mechanism mainly suppresses large tracking errors, motivating the use of dispersion statistics in addition to RMSE.

\subsubsection{Measurement gating and association}
Following ByteTrack~\cite{zhang2022bytetrack}, we use $1-\mathrm{IoU}$ as the association cost and the Hungarian algorithm~\cite{weng2020ab3dmot} for one-to-one matching. To handle panorama seam crossings, the IoU is evaluated under horizontal shifts of $-W$, $0$, and $+W$, and the maximum is used. Unmatched high-confidence detections initialize new tracks, while unmatched tracks are retained for a short patience window.

After association, matched measurements are validated by a $\chi^2$ NIS gate, with thresholds of $8.0$ for scalar updates and $15.0$ for 5-D joint updates. If a joint update is rejected, the filter falls back to the image-only update to avoid the influence of inconsistent depth measurements.

\subsubsection{Noise modeling and numerical safeguards}
The joint measurement covariance $\mathbf{R}$ is block diagonal except for the $h$ and $d$ covariance block in~\eqref{eq:cov_hd}. For the directional measurement, a 3-pixel detection uncertainty is mapped to the spherical domain through the panoramic geometry, with $\sigma_{\mathrm{dir}}$ scaled by the target angular height $\alpha$ to account for the target-size-dependent uncertainty of panoramic detection. The depth variance accounts for LiDAR range noise, the standard error of the cluster median, and uncertainty in body-surface selection, and is capped at $0.25$\,m$^{2}$. The process noise is defined separately for different state components. The effective meridional extent and distance channels follow the standard continuous-white-noise-acceleration (CWNA) model, while the direction and angular aspect-ratio channels use a diagonal random-walk model, $Q_{ii}=\sigma_i^2\Delta t$, with $\sigma_i\propto\alpha$ \reviewed{for the direction channels}. All parameter values are listed in \reviewed{Table~VIII of Appendix~B}.

For numerical stability, we clamp small eigenvalues of $\mathbf{S}_{k,m}$, symmetrize $\mathbf{P}_{k|k}$, and impose a diagonal lower bound of $10^{-4}$ on $\mathbf{P}_{k|k}$. For small angular displacements ($\|\mathbf{w}_k\|\to0$), the first-order exp-map approximation in Appendix~A is used to avoid numerical instability.

\subsubsection{Algorithm}

\begin{algorithm}[t]
\caption{Extended Spherical Kalman Filter for Panoramic MOT}
\label{alg:extendedKF}
\begin{algorithmic}[1]
\Require Previous $(\hat{\mathbf{x}}_{k-1|k-1},\mathbf{P}_{k-1|k-1},\bar{\mathbf{g}}_{k-1},\mathbf{B}_{k-1})$;
panoramic detections $\{(u,v,s_x,s_y,\text{score})\}$; synchronized LiDAR scan; \reviewed{time step} $\Delta t$
\State \textbf{Predict} $\hat{\mathbf{x}}_{k|k-1}=\mathbf{F}\hat{\mathbf{x}}_{k-1|k-1}$,\;
$\mathbf{P}_{k|k-1}=\mathbf{F}\mathbf{P}_{k-1|k-1}\mathbf{F}^\top+\mathbf{Q}$; advance $\bar{\mathbf{g}}$ by the exp-map, rebuild $\mathbf{B}$, reset $(w_1,w_2)\leftarrow 0$
\State Pair the frame with the nearest deskewed scan ($\le 60$\,ms), splat it to the panoramic depth map, and extract a per-box cluster depth $(d_k^{\mathrm{obs}},R_{d})$ (nearest dense cluster; none if the cluster test fails)
\State For each detection, compute the log-map direction $\mathbf{z}_{\text{dir},k}=[z_{1,k},z_{2,k}]^\top$, the angular height $\alpha_k^{\text{det}}$ via \eqref{eq:alpha_det}, and the angular aspect ratio $\ell_k^{\text{det}}$ via \eqref{eq:ell_det}
\State \textbf{Associate} detections to tracks by seam-aware image-plane IoU cost and Hungarian matching~\cite{zhang2022bytetrack,weng2020ab3dmot}; predicted boxes are generated by the spherical EKF prediction
\State \textbf{Update} each matched track using the joint 5-D measurement \eqref{eq:z_joint}--\eqref{eq:cov_hd} when depth is available and passes the NIS gate; otherwise, use the image-only update \eqref{eq:z_img}--\eqref{eq:jac_alpha}
\State Handle unmatched tracks and initialize new tracks from unmatched high-confidence detections, using measured depth when available and the nominal scale prior otherwise, following~\cite{zhang2022bytetrack}
\State Reproject to sphere: $\bar{\mathbf{g}}_k\!=\!\mathrm{Exp}_{\bar{\mathbf{g}}^{\text{pred}}_k}\!\big(\hat{\mathbf{w}}_{k|k}\big)$; normalize; update basis $\mathbf{B}_k$; reset \reviewed{$(w_1,w_2)\leftarrow 0$}
\State Parallel-transport velocities with $\mathbf{T}_k=\mathbf{B}_k^\top\mathbf{B}^{\text{tmp}}_k$
\State \Return Updated $(\hat{\mathbf{x}}_{k|k},\mathbf{P}_{k|k},\bar{\mathbf{g}}_k,\mathbf{B}_k)$ and the tracked box $(u,v,s_x,s_y)$ with range $d_k=x_{9}$
\end{algorithmic}
\end{algorithm}

This pipeline (Algorithm~\ref{alg:extendedKF}) preserves spherical geometry for direction estimation (avoiding image-space singularities~\cite{sun2024smmpod,geyer2001ijcv}), incorporates ByteTrack-style box dynamics~\cite{zhang2022bytetrack}, and fuses rotating-LiDAR depth to stabilize association and metric scale in panoramic scenes.

\section{Map-matching based Ground Truth Generation for 3D Tracking}

\subsection{Hardware and System Configuration}
\begin{figure}[]
\centering
    \includegraphics[width=\linewidth]{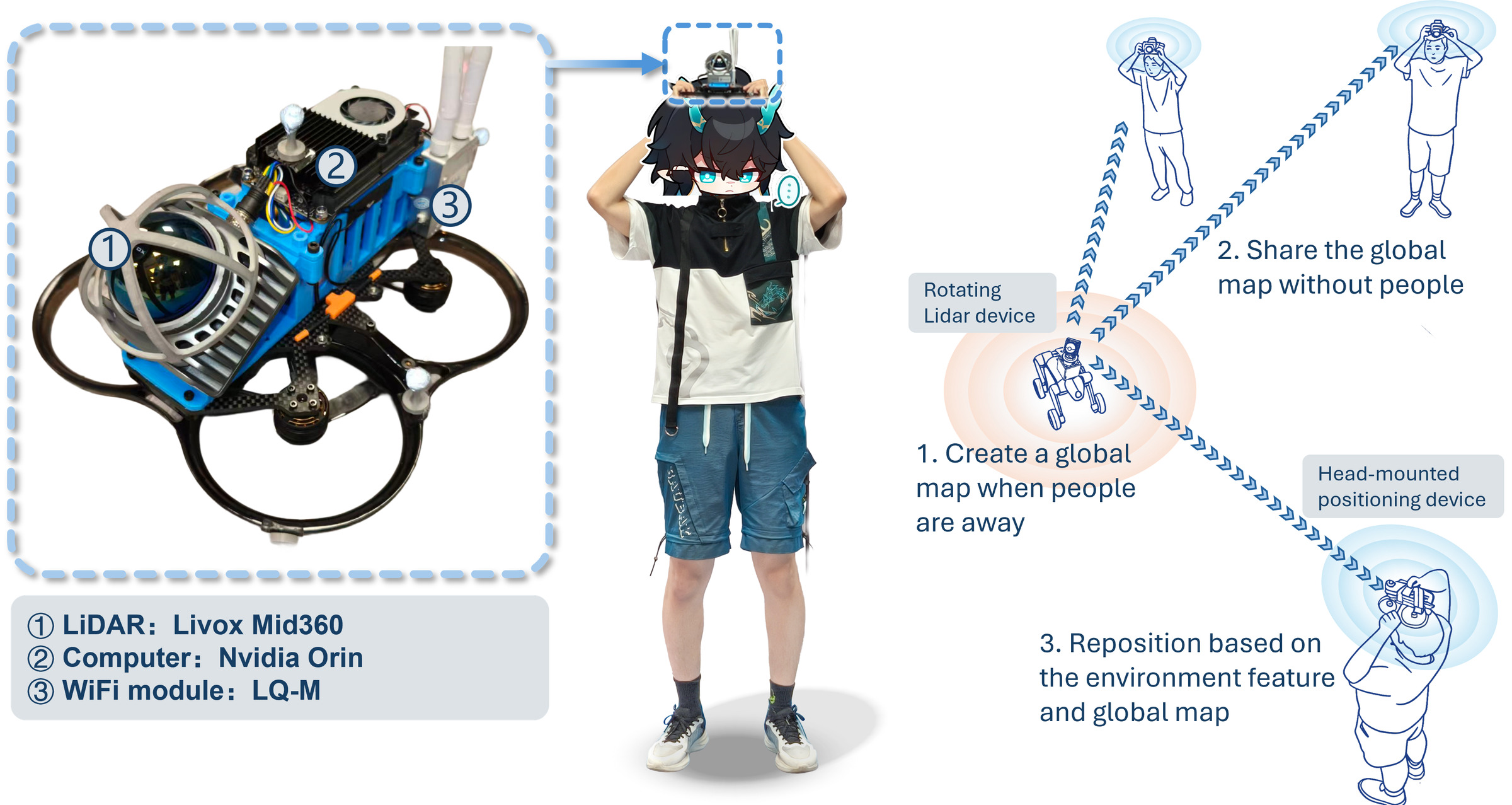}
    \caption{The hardware of the localization system for each agent. Each person is wearing a LiDAR and a wireless module, which will be used to generate \reviewed{a ground-truth trajectory}.}
    \label{fig:localization_system}
\end{figure}

To enable rapid human localization in open outdoor environments, we design a mobile localization system that requires no pre-deployed base stations. The system consists of an arbitrary number of wearable localization devices (as shown on the left of Fig.~\ref{fig:localization_system}) and a WiFi router; each device integrates a LiDAR sensor and a compact onboard computing unit, providing independent perception and computation capabilities.

During operation, the device is worn on top of a person's head (as shown in the middle of Fig.~\ref{fig:localization_system}), continuously perceives the environment, and performs self-localization against a pre-built global map, thereby obtaining the position of the human carrier. All devices are connected to the same WiFi network for data communication and time synchronization. Concretely, all devices are time-synchronized via NTP over this network. The residual content-time offsets between the camera and the rotating LiDAR ($\approx 0.15$\,s), as well as those between the ground-truth devices and the recording sessions, are calibrated once per session and compensated, as detailed in Section~V-B.

% \subsection{3D Tracking Data Generation}

% Before conducting the experiments, a unified global reference frame was established by constructing a global map using a rotating LiDAR device. With all participants kept away from the experimental area, the LiDAR sensor was manually rotated and moved while running the FAST-LIO algorithm to perform environment mapping. After sufficient motion excitation, the sensor remained stationary for a short period to improve optimization convergence and map stability, resulting in a high-quality global point cloud map.

% Once the mapping process was completed, the generated global map was distributed to all localization devices as a reference for localization. Each device then performed LiDAR-based localization against the pre-built map to estimate its pose within the shared global coordinate frame. After successful initialization, the localization device was mounted on the experimenter’s head to enable real-time acquisition of the human 3D trajectory.

% Since the global map shares the same coordinate frame as the LiDAR device used during the mapping stage, all localization devices and experimental sensing platforms operate within a unified global coordinate system. This ensures spatial consistency across multiple devices and enables reliable generation of 3D tracking data for subsequent analysis.

% Furthermore, separating the mapping and localization stages avoids repeated mapping during experiments, improves experimental efficiency, and enhances repeatability and consistency across different experimental trials.

\subsection{3D Tracking Ground-truth Generation}

To generate reliable 3D tracking ground truth without motion-capture infrastructure, we establish a unified global coordinate frame based on a high-quality LiDAR point cloud map as shown on the right of Fig.~\ref{fig:localization_system}: our rotating-LiDAR system first constructs a global map of the experimental environment using a tightly coupled LiDAR--inertial odometry pipeline \cite{li2025graph}, with the sensing area kept free of dynamic objects during mapping.

All wearable localization devices subsequently perform online localization by registering their real-time LiDAR scans to this pre-built map, yielding full 6-DoF poses per device in the shared map coordinate system. In other words, a single global coordinate frame is used throughout: the rotating-LiDAR platform builds the map, and each wearable device performs FAST-LIO-based relocalization~\cite{xu2022fast} against this shared map and directly outputs its 6-DoF pose in the global frame. The ground-truth and tracker trajectories are therefore expressed in the same frame, with no inter-device calibration or cross-frame transformation.

The system does not impose a fixed maximum operating range, since the coverage can be extended by enlarging the pre-built map. Its planar localization accuracy is within 3\,cm, based on LiDAR--inertial odometry and relocalization methods~\cite{xu2022fast}, which is substantially smaller than the tracking errors reported in Section~V. Although this localization accuracy was not independently re-evaluated under dynamic human-worn conditions in this study, it provides a sufficiently accurate reference for quantitative tracking evaluation.

By separating the mapping and localization stages, the pipeline avoids cumulative drift during experiments, improves repeatability across trials, and enables scalable multi-person trajectory acquisition without infrastructure constraints.

\section{Experiments}

\subsection{Implementation Details}

All experiments were conducted onboard an NVIDIA Jetson AGX Orin embedded platform
running Ubuntu~20.04 with ROS~1. The system was implemented in C++ using Eigen
for numerical computation and OpenCV for image processing.

The four fisheye images are projected into a $1280\times640$ equirectangular panorama with a vertical field of view of $\pm90^{\circ}$. Person detection is performed using the COCO-pretrained YOLO11-m model without in-domain fine-tuning. The extended spherical Kalman filter and data association run at the camera frame rate of $15$\,Hz. Additional implementation details and parameter settings are provided in \reviewed{Table~VIII of Appendix~B}.

Table~\ref{tab:runtime} summarizes the runtime of each processing stage on the onboard platform. The complete pipeline operates within the $15.1$\,Hz sensor budget, with a median per-frame latency of approximately $55.0$\,ms and no frame drops in the profiled sequence. Panoramic projection accounts for most of the computation, while the tracking stage requires only $2.4$\,ms per frame.

\begin{table}[t]
\centering
\caption{Per-stage runtime on the Jetson AGX Orin (559 frames).}
\label{tab:runtime}
\small

\resizebox{\columnwidth}{!}{%
\begin{tabular}{lccc}
\toprule
Stage & Median (ms) & P95 (ms) & Max (ms) \\
\midrule
Panoramic projection (CPU) & 35.0 & 37.0 & 66.7 \\
Detection (TensorRT FP16) & 17.6 & 20.9 & -- \\
Tracking (EKF + association) & 2.4 & 3.3 & 11.0 \\
\midrule
Serial total & $\approx$55.0 & $\approx$61 & -- \\
\bottomrule
\end{tabular}}

\end{table}

\subsection{Evaluation Metrics and Protocol}

To quantitatively evaluate tracking accuracy, we compare the estimated human
trajectories with reference trajectories obtained from the global-map-based
localization system described in the previous section.

\textbf{Reference-point alignment.} The wearable localization device is mounted on top of each participant's head, whereas the tracker estimates the body-center position. For consistent 3D evaluation, we reconstruct the corresponding head-top position for each tracked target using the bearing of the midpoint of the upper edge of the detection box and the LiDAR depth estimated at the body center. This reconstructed point corresponds to the mounting location of the wearable device, allowing direct comparison with the ground-truth trajectory without additional head-to-body offset compensation.

\textbf{Coordinate and time alignment.} All estimates and reference trajectories are expressed in the single global LiDAR map frame (Section~IV-B). The reference trajectories (10\,Hz) are linearly interpolated to the tracker timestamps (15.1\,Hz) with a nearest-sample tolerance of 50\,ms (half the sampling interval); no extrapolation across gaps is performed. Timestamps across devices are aligned via NTP, and the residual content-time offsets described in Section~IV-A are calibrated per session and compensated.

\textbf{Metrics.} We report (i) the 3D RMSE between the head-top estimates and the reference trajectories, together with its radial, tangential, and vertical components relative to the sensor position. The radial direction follows the horizontal line of sight, while the tangential direction lies in the horizontal plane and is perpendicular to it. We also report (ii) standard MOT metrics, including MOTA, MOTP, IDF1, HOTA, FP, FN, IDS, fragmentation, and MT/PT/ML. MOTA, MOTP, IDF1, FP, FN, IDS, fragmentation, and MT/PT/ML are computed with py-motmetrics, while HOTA and its detection (DetA) and association (AssA) components follow the formulation of Luiten et al.~\cite{luiten2021hota}; \reviewed{the main per-sequence and merged metrics are summarized in Table~\ref{tab:mot}, and the complete metrics are reported in Table~IX of Appendix~B.} Since the ground truth consists of 3D points rather than bounding boxes, identity association is performed in the global map frame using 3D Euclidean distance with a matching threshold of $d_{\max}=1.0$\,m. For HOTA, the similarity is defined as $S=\max(0,1-d/d_{\max})$. Only tracking outputs with valid 3D positions are included in the evaluation. The five sequences are evaluated independently, and identities are offset across sequences when computing the merged statistics to avoid identity conflicts.

\subsection{Evaluation in Real-world Environment}

% \begin{figure*}[t]
% \includegraphics[width=\linewidth]{figs/tracking_exp.png}
% \caption{Real-time tracking results in the experimental environment.
% (a)--(d) show four representative moments.
% Top row: depth projections on the panoramic images.
% Bottom row: reconstructed 3D trajectories of multiple tracked persons.}
% \label{fig:tracking_exp}
% \end{figure*}

We evaluate the proposed tracker in a real-world outdoor experimental environment
with three participants moving freely within the sensing area.
Each participant wears a localization device described previously,
allowing synchronized acquisition of reference trajectories and tracking results.

\begin{figure}[]
    \centering
\includegraphics[width=1.0\linewidth]{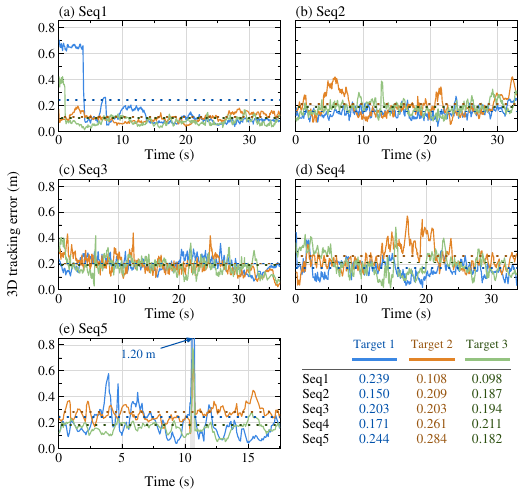}
    \caption{3D tracking error \reviewed{over time for the three targets in (a)--(e) Seq1--Seq5 on a common axis; dotted lines mark the per-target RMSE (m), also listed in the inset table.}}
    \label{fig:tracking_error}
\end{figure}

Following the evaluation protocol of Section~V-B, we report the full 3D errors and standard MOT metrics. The detailed 3D tracking errors for the five sequences
(Seq1--Seq5) are summarized in Table~\ref{tab:err3d}.
The error curves over time are shown
in Fig.~\ref{fig:tracking_error}.

Across all trials, the proposed method achieves consistent decimeter-level tracking accuracy: the merged 3D RMSE is 0.197\,m over 6845 evaluated frames, and the per-target 3D RMSE ranges from 0.098 to 0.284\,m over the 15 tracked targets.
The five sequences cover complementary motion conditions (Table~\ref{tab:err3d}, Motion column): static (Seq1), clockwise and counterclockwise circumnavigation (Seq2--Seq3), faster counter-directed motion with frequent mutual occlusions (Seq4), and fast radial back-and-forth motion producing rapid depth variation (Seq5). Static Seq1 achieves the smallest error, while the larger errors in the dynamic sequences stem from faster motion, occlusions, and the inherent processing latency, which increases temporal misalignment between sensing and state estimation.
No explicit latency compensation is applied; tighter coupling with low-level rotating-LiDAR odometry to reduce this misalignment is listed as future work.
Overall, the system maintains stable multi-person tracking across static, dynamic, fast-motion, and occlusion-rich scenarios, with the wrap-continuous spherical state and the depth-fused filter suppressing noise and preserving trajectory consistency during complex human motions.

Table~\ref{tab:err3d} summarizes the per-sequence 3D error decomposition using the head-top-aligned evaluation protocol of Section~V-B, with per-target trajectory pairing. Across all 6845 evaluated frames, the merged 3D RMSE is 0.197\,m. The vertical component has an RMSE of 0.088\,m with a bias of $+0.064$\,m, while the horizontal-plane RMSE is 0.176\,m. Overall, the vertical component accounts for 20.0\% of the total squared error.

\begin{table}[t]
\centering
\caption{3D error decomposition per sequence (m): bias/RMSE, using per-target trajectory pairing under the head-top-aligned protocol of Section~V-B. The motion attribute of each sequence is indicated. (The per-frame nearest-neighbor pairing with a 5\,m gate used in Table~\ref{tab:ablation} yields a merged RMSE of 0.276\,m.)}
\label{tab:err3d}

\resizebox{\columnwidth}{!}{%
\begin{tabular}{lccccc}
\toprule
Seq. & Motion & Radial & Tangential & Vertical & 3D RMSE \\
\midrule
Seq1 & Static & $-0.060$/0.142 & $-0.014$/0.043 & $+0.039$/0.064 & 0.161 \\
Seq2 & CW & $-0.122$/0.150 & $+0.016$/0.075 & $+0.063$/0.075 & 0.183 \\
Seq3 & CCW & $-0.083$/0.119 & $-0.093$/0.122 & $+0.092$/0.105 & 0.200 \\
Seq4 & Fast + occl. & $-0.108$/0.158 & $+0.066$/0.127 & $+0.043$/0.077 & 0.217 \\
Seq5 & Fast radial & $-0.033$/0.139 & $-0.073$/0.149 & $+0.101$/0.128 & 0.241 \\
\midrule
All & -- & $-0.086$/0.142 & $-0.016$/0.105 & $+0.064$/0.088 & \textbf{0.197} \\
\bottomrule
\end{tabular}}

\end{table}

Table~\ref{tab:mot} summarizes the standard MOT metrics of the proposed method and the image-plane Kalman baseline. Our method outperforms the baseline in the merged results (MOTA $0.983$ vs.\ $0.860$; zero identity switches vs.\ $20$), and the gap widens precisely where depth observations are sparse: valid LiDAR depth is available for only $34$--$43\%$ of the frames per target, with blind intervals up to $9.8$--$12.3$\,s, and in Seq3 (the longest depth gaps) the baseline MOTA drops to $0.710$ versus $0.997$ for ours. Our method also produces no identity switches across the 16 ground-truth seam crossings (vs.\ 20 for the baseline), thanks to the wrap-continuous angular state and consistent association across the image boundary. The baseline reaches a slightly higher HOTA in Seq1 ($0.922$ vs.\ $0.915$), while our method is better in the remaining sequences and in the merged results.

Overall, these results demonstrate improved tracking robustness under intermittent depth observations and panorama seam crossings.

\begin{table}[t]
\centering
\caption{MOT metrics per sequence ($d_{\max}=1.0$\,m).}
\label{tab:mot}
\small

\resizebox{\columnwidth}{!}{%
\begin{tabular}{llccccc}
\toprule
Seq. & Method & MOTA & MOTP & IDF1 & HOTA & IDS \\
\midrule
\multirow{2}{*}{Seq1} & Ours & 0.997 & 0.114 & 0.999 & 0.915 & 0 \\
 & Baseline & 0.952 & 0.087 & 0.976 & 0.922 & 0 \\
\multirow{2}{*}{Seq2} & Ours & 0.989 & 0.173 & 0.994 & 0.854 & 0 \\
 & Baseline & 0.907 & 0.260 & 0.540 & 0.520 & 5 \\
\multirow{2}{*}{Seq3} & Ours & 0.997 & 0.190 & 0.998 & 0.844 & 0 \\
 & Baseline & 0.710 & 0.311 & 0.439 & 0.420 & 8 \\
\multirow{2}{*}{Seq4} & Ours & 0.944 & 0.199 & 0.972 & 0.817 & 0 \\
 & Baseline & 0.898 & 0.224 & 0.498 & 0.530 & 6 \\
\multirow{2}{*}{Seq5} & Ours & 0.991 & 0.211 & 0.995 & 0.822 & 0 \\
 & Baseline & 0.827 & 0.313 & 0.904 & 0.671 & 1 \\
\midrule
\multirow{2}{*}{All} & Ours & \textbf{0.983} & \textbf{0.173} & \textbf{0.992} & \textbf{0.856} & \textbf{0} \\
 & Baseline & 0.860 & 0.227 & 0.649 & 0.623 & 20 \\
\bottomrule
\end{tabular}}
\end{table}

\subsection{Deployment on Aerial and Ground Platforms}

\begin{figure}[t]
\centering
\includegraphics[width=\linewidth]{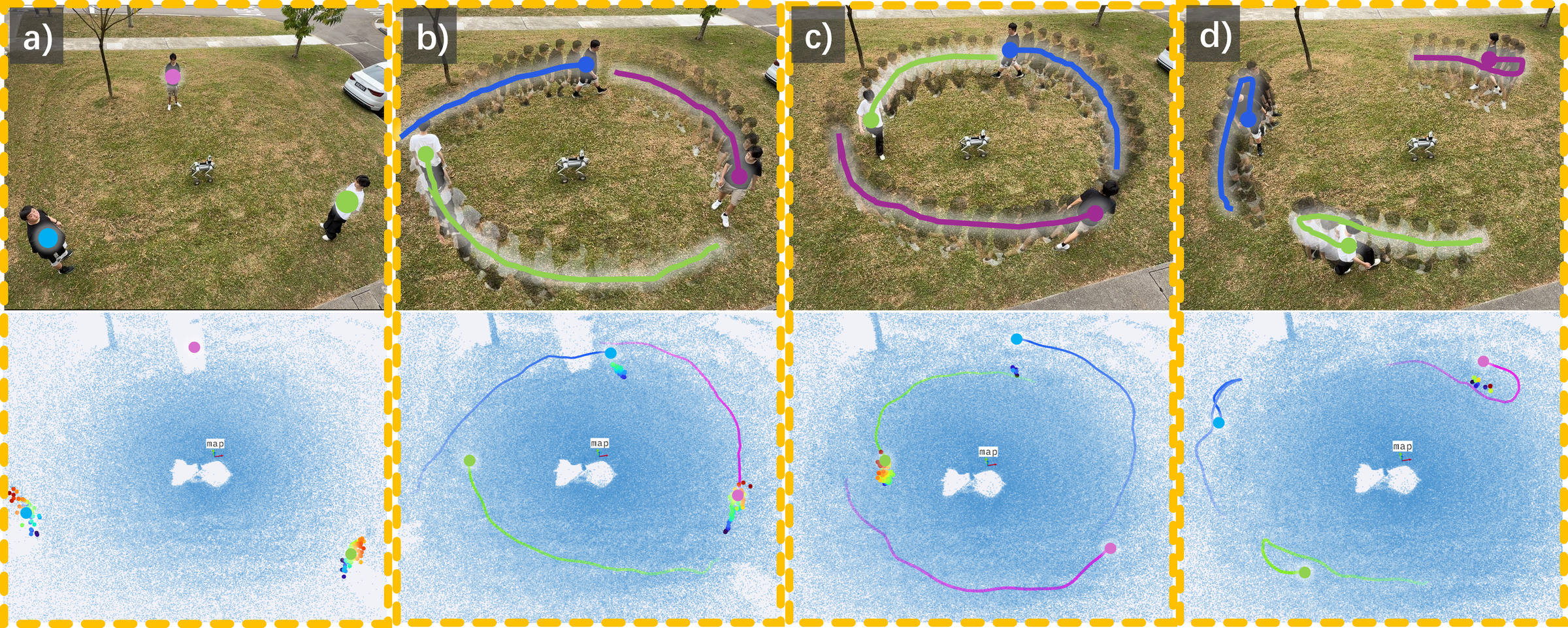}
    \caption{Evaluation on the robot dog in a real-world outdoor environment. (a) Objects standing. (b) Objects moving clockwise. (c) Objects moving counterclockwise. (d) Objects suddenly changing direction.}
    \label{fig:outdoor_dog_test}
\end{figure}

\begin{figure}[t]
\centering
\includegraphics[width=\linewidth]{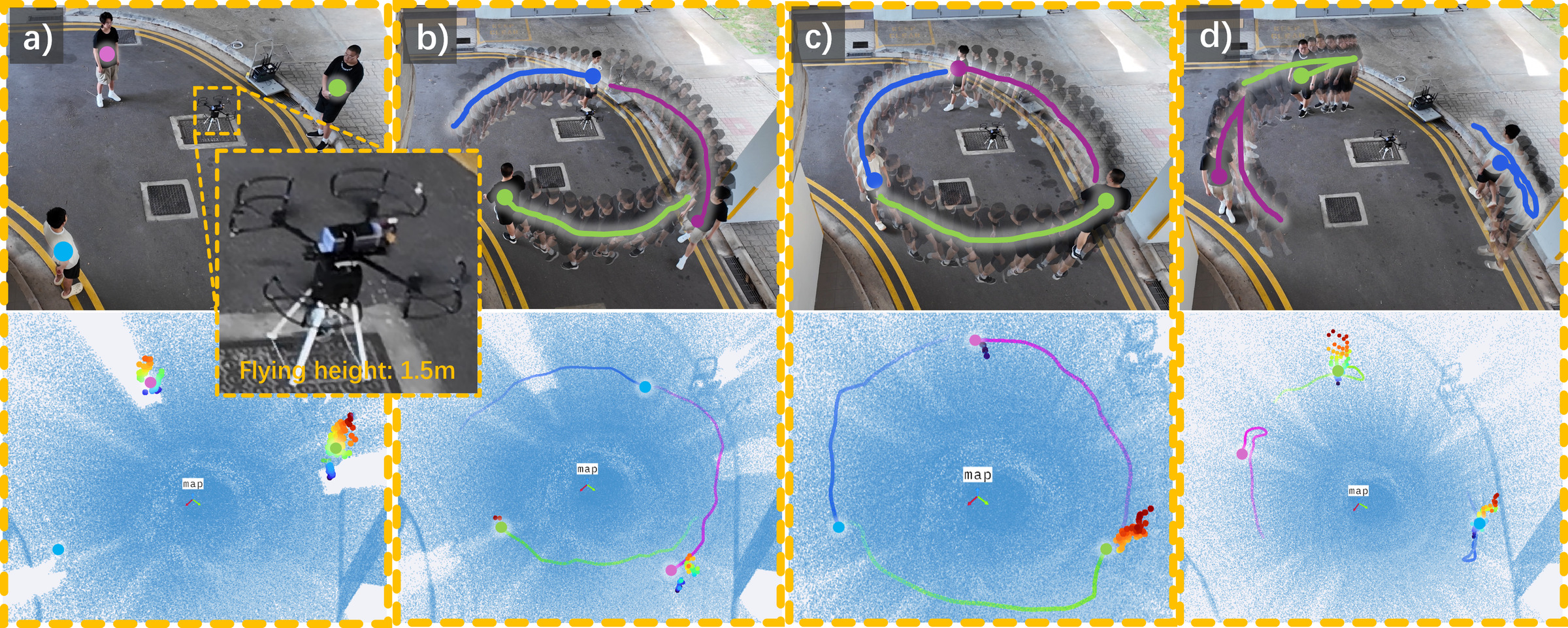}
    \caption{Evaluation on the drone in a real-world outdoor environment. (a) Objects standing. (b) Objects moving clockwise. (c) Objects moving counterclockwise. (d) Objects suddenly changing direction.}
    \label{fig:outdoor_drone_test}
\end{figure}

% To further evaluate the robustness and generalization capability
% of the proposed tracking framework, we conducted experiments
% in diverse real-world scenarios, as shown in Fig.\ref{fig:outdoor_dog_test}.
% The experiments were performed on multiple mobile platforms.
% Specifically, the sensing system was mounted on a quadruped robot
% and tested in both indoor and outdoor environments.

% In addition, the same system was deployed on an unmanned aerial
% vehicle (UAV), where human tracking experiments were carried out
% under indoor and outdoor flight conditions.

% The wearable LiDAR localization devices introduced previously
% were used only to provide reference trajectories when available.
% However, in large-scale real-world environments, accurate
% ground-truth trajectories are difficult to obtain.
% Therefore, evaluation is primarily conducted qualitatively
% through visual inspection of reconstructed 3D trajectories
% projected onto panoramic imagery and LiDAR point clouds. 
% Across all scenes, our tracker operates in real time on the onboard AGX Orin  platform and produces stable trajectories for pedestrians, cyclists, and vehicles. 
% The estimated 3D tracks remain consistent, demonstrating the robustness of the spherical state-space representation and multimodal fusion in unconstrained environments. 
% These results confirm that the proposed system is not only effective in controlled laboratory settings but also applicable to practical outdoor and indoor deployments.

To assess cross-platform generalization, we deploy the panoramic sensing system on a quadruped robot (Fig.~\ref{fig:outdoor_dog_test}) and a UAV (Fig.~\ref{fig:outdoor_drone_test}), demonstrating that the proposed formulation operates on both platforms tested in this work.

\textbf{Ground deployment.}
The sensing module was mounted on a quadruped robot and evaluated in both indoor and outdoor environments with static and moving targets, including pedestrians and cyclists. Despite ego-motion disturbances and rapid viewpoint changes, the tracker maintained stable identities and smooth 3D trajectories, with full-surround coverage even when targets moved behind the robot or across large angular spans.

\textbf{Aerial deployment.}
The same system was integrated onto a UAV and tested under hovering and low-altitude flight conditions (altitude $\approx$1.3--1.7\,m above ground). Aerial motion adds vibration, altitude variation, and larger perspective changes, yet the spherical filter preserved directional consistency on $\mathbb{S}^2$ and maintained stable multimodal fusion under rapid ego-motion.

Across all tested sequences, no evident tracking loss, divergence, or oscillatory behavior was observed in our experiments: identities remained stable through long-duration recordings even across large angular spans, temporary field-of-view exits, and partial occlusions, with the predicted motion model allowing tracks to persist and re-associate correctly once the targets reappeared; the estimated trajectories were smooth and consistent with visual observations.

\subsection{Ablation Study and Baseline Comparison}
\label{sec:ablation}

To evaluate the contribution of each design component, all ablation variants and baselines use the same per-frame detections and associated LiDAR depth obtained before tracking. The detector, panoramic projection, depth extraction, back-projection geometry, and evaluation protocol are kept unchanged, with each ablation variant modifying only the component under evaluation. Table~\ref{tab:ablation} reports the 3D error statistics and MOT metrics defined in Section~V-B for each variant.\footnote{The error statistics in Table~\ref{tab:ablation} use per-frame nearest-neighbor pairing within a 5\,m gating radius, ensuring that drifted estimates are retained as errors rather than excluded (coverage $\approx100\%$). The MOT metrics use $d_{\max}=1.0$\,m. This pairing rule differs from that used in Table~\ref{tab:err3d}, which assumes intact trajectories, whereas the baseline methods may produce fragmented tracks that do not support the same global identity assignment.} Table~\ref{tab:baseline} provides further comparisons with stronger baseline trackers.

The variants are: (i) \emph{w/o depth fusion}---the LiDAR depth never arrives and only image updates run; (ii) \emph{Cartesian EKF (height-coupled)}---the position is re-parameterized in Cartesian coordinates in the sensor frame with the same state information (bearing, angular height, intermittent range) and the same height-coupling and gain-decay mechanisms as the full method, so that the comparison isolates the parameterization as the main design difference; (iii) \emph{w/o seam-aware IoU}---the $\pm W$ wrap-around is disabled in the association cost on the method side (measured on the Cartesian twin, which is metrically indistinguishable from the full method and whose seam-aware IoU is bit-identical to the C++ implementation); (iv) \emph{baseline + seam-aware IoU}---the wrap-around is instead added to the image-plane baseline; (v) \emph{w/o blind-zone gain decay}---the blind-zone weight is set to $w\equiv 1$ with no height-velocity freezing; and (vi) the \emph{2D image-plane KF} baseline (ByteTrack-style pixel state; the per-box depth is zero-order-held for output but never enters the filter).

\begin{table*}[!t]
\centering
\caption{Ablation on the five sequences (merged, $n=6894$ paired frames). Error statistics use a 5\,m gate with per-frame nearest-neighbor pairing (drift counted as error); MOT metrics use $d_{\max}=1.0$\,m.}
\label{tab:ablation}
\small

\resizebox{\textwidth}{!}{%
\begin{tabular}{lccccc|ccccc}
\toprule
Variant & 3D RMSE (m) & std & P95 & P99 & f$>$0.4 & MOTA & IDF1 & HOTA & AssA & IDS \\
\midrule
Full S$^3$KF & \textbf{0.276} & \textbf{0.204} & \textbf{0.339} & \textbf{0.661} & \textbf{2.8\%} & \textbf{0.983} & \textbf{0.992} & \textbf{0.856} & \textbf{1.000} & \textbf{0} \\
w/o depth fusion & 0.381 & 0.226 & 0.645 & 0.734 & 19.4\% & 0.981 & 0.976 & 0.735 & 0.980 & 1 \\
Cartesian EKF (height-coupled) & 0.279 & 0.204 & 0.344 & 0.667 & 3.2\% & 0.983 & 0.992 & 0.853 & 1.000 & 0 \\
w/o seam-aware IoU & 0.302 & 0.230 & 0.348 & 0.749 & 3.2\% & 0.977 & 0.911 & 0.809 & 0.911 & 17 \\
w/o blind-zone gain decay & 0.283 & 0.208 & 0.389 & 0.643 & 4.6\% & 0.983 & 0.992 & 0.851 & 1.000 & 0 \\
2D image-plane KF baseline & 0.825 & 0.705 & 2.652 & 3.160 & 20.8\% & 0.860 & 0.649 & 0.623 & 0.697 & 20 \\
\quad + seam-aware IoU & 0.823 & 0.703 & 2.651 & 3.155 & 20.7\% & 0.864 & 0.653 & 0.626 & 0.700 & 15 \\
\bottomrule
\end{tabular}}

\end{table*}

Three observations can be made from the ablation results. First, LiDAR depth fusion provides the largest improvement in tracking accuracy. Removing depth fusion increases the 3D RMSE from 0.276\,m to 0.381\,m ($+38\%$), increases the fraction of errors above 0.4\,m from 2.8\% to 19.4\%, and reduces HOTA from 0.856 to 0.735. Second, without LiDAR depth, the proposed method still substantially outperforms the image-plane baseline, reducing the 3D RMSE from 0.825\,m to 0.381\,m and improving HOTA from 0.623 to 0.735. This indicates that the wrap-continuous angular state and angular-height coupling remain effective when depth is unavailable. Third, the Cartesian counterpart performs similarly to the full method, with 3D RMSEs of 0.279\,m and 0.276\,m, respectively, and identical MOTA, IDF1, and identity metrics; the residual difference is small (0.003\,m, $1.1\%$ of the full method's RMSE). These results indicate that the main performance gains arise from joint depth fusion, angular-height coupling during image-only intervals, and blind-zone gain decay, rather than from the spherical parameterization alone.

The gain decay mainly suppresses large errors rather than the error level: disabling it leaves the median error and the MOT metrics unchanged but raises P95 from 0.339\,m to 0.389\,m and the above-0.4\,m fraction from 2.8\% to 4.6\%, which motivates reporting dispersion statistics in addition to RMSE. Seam-aware association and the wrap-continuous state are complementary: disabling the wrap-around on our side increases identity switches from 0 to 17 (IDF1 0.992$\to$0.911, HOTA 0.856$\to$0.809) and the 3D RMSE by about 9\%, whereas adding it to the image-plane baseline only reduces its identity switches from 20 to 15, because with a pixel-based state a corrected association still admits a full-panorama-width innovation across the image boundary.

\begin{table}[t]
\centering
\caption{Comparison with stronger baselines (merged over the five sequences; protocol as in Table~\ref{tab:ablation}).}
\label{tab:baseline}

\resizebox{\columnwidth}{!}{%
\begin{tabular}{lccccccc}
\toprule
Method & 3D RMSE (m) & P95 & f$>$0.4 & MOTA & IDF1 & AssA & IDS \\
\midrule
Full S$^3$KF & \textbf{0.276} & \textbf{0.339} & \textbf{2.8\%} & \textbf{0.983} & \textbf{0.992} & \textbf{1.000} & \textbf{0} \\
2D image-plane KF & 0.825 & 2.652 & 20.8\% & 0.860 & 0.649 & 0.697 & 20 \\
Cartesian 3D KF & 0.612 & 1.316 & 23.3\% & 0.841 & 0.729 & 0.777 & 24 \\
\bottomrule
\end{tabular}}

\end{table}

Table~\ref{tab:baseline} includes a stronger baseline, a Cartesian \reviewed{constant-velocity} Kalman tracker in the map frame following AB3DMOT~\cite{weng2020ab3dmot}, independently reimplemented using the same detections. Compared with the \reviewed{image-plane} baseline, it reduces the 3D RMSE from 0.825 to 0.612\,m and improves IDF1 from 0.649 to 0.729 and AssA from 0.697 to 0.777. The remaining error is mainly radial, with a standard deviation of 0.455\,m compared with 0.182\,m for our full method. The \reviewed{six-dimensional} $[\mathbf{p},\mathbf{v}]$ state lacks a \reviewed{height-related} constraint, making distance difficult to maintain when LiDAR depth is unavailable, which occurs in 62.3\% of the detection frames. We also swept the process noise over $q\in[0.01,100]$ \reviewed{(Fig.~9 of Appendix~B)}, with the best overall result at $q=1.0$. For pooled cart\_map, the 1 m-gated 3D RMSE varies by only 0.017 m over \(q\in[1,30]\), indicating low sensitivity to \(q\). Similar sweeps of the \reviewed{image-plane} baseline, including track buffer duration and detector thresholds, produced nearly identical results.

Evaluation on KITTI or nuScenes is not directly applicable because their standard 3D tracking pipelines rely on oriented 3D boxes from automotive LiDAR detectors, which are not available with our rotating panoramic sensing setup. We therefore leave evaluation on public benchmarks to future work.

\subsection{Discussion}
The core novelty of this work is the geometry-consistent state formulation directly on the unit sphere $\mathbb{S}^2$: object bearings use a two-degree-of-freedom tangent-plane model, which preserves intrinsic spherical geometry while enabling linear-Gaussian inference in a locally Euclidean space.
The ablation study in Section~\ref{sec:ablation} shows that the parameterization itself is not the main source of the accuracy and identity improvements. A Cartesian EKF with the same state information differs from the full S$^3$KF by only 0.003\,m in 3D RMSE (1.1\%), while achieving identical MOTA, IDF1, and identity metrics. The main gains instead come from two aspects. First, the continuous angular state avoids the discontinuity of \reviewed{pixel-based} states at the panorama seam. Second, the fusion mechanisms improve tracking under intermittent depth observations by incorporating \reviewed{per-frame} depth as a joint observation with scale and depth covariance, coupling angular measurements with object height to stabilize the distance estimate during depth outages, and applying gain decay in depth blind zones. The \reviewed{tangent-plane} parameterization provides a natural way to maintain angular continuity across the panorama seam.

Compared with 2D trackers such as ByteTrack~\cite{zhang2022bytetrack}, whose pixel-coordinate states suffer projection singularities in panoramic imagery, direction estimation here remains numerically stable across the full $360^{\circ}$ field of view.
The operative property is wrap-continuity: a pixel-coordinate state jumps by the full panorama width at the left--right wrap-around boundary, corrupting the filter state even when association succeeds, whereas the tangent-plane coordinates vary continuously through the crossing.

To further quantify identity stability on four additional self-collected panoramic sequences without ground-truth annotations, we compare the \emph{maximum assigned track ID} per sequence---a proxy for track fragmentation and identity proliferation, since a tracker that frequently fragments and re-assigns identities accumulates larger track IDs---between the proposed tracker and ByteTrack. Each sequence contains three participants moving around the sensing platform, \reviewed{and the four sequences cover} both platforms and environments: \textit{Ground-Outdoor}, \textit{Ground-Indoor}, \textit{Aerial-Outdoor}, and \textit{Aerial-Indoor} (Table~\ref{tab:maxid_compare}).

\begin{table}[h]
\centering
\caption{Track-ID proliferation on four additional panoramic sequences: maximum assigned track ID per sequence.}
\label{tab:maxid_compare}
\begin{tabular}{lcc}
\toprule
Sequence & ByteTrack (2D) & Proposed tracker \\
\midrule
Ground-Outdoor & 26 & 3 \\
Ground-Indoor & 18 & 3 \\
Aerial-Outdoor & 27 & 3 \\
Aerial-Indoor & 27 & 7 \\
\bottomrule
\end{tabular}
\end{table}

Table~\ref{tab:maxid_compare} shows that the sum of the per-sequence maximum assigned track IDs decreases from 98 (ByteTrack) to 16, dropping to only three in \textit{Ground-Outdoor}, \textit{Ground-Indoor}, and \textit{Aerial-Outdoor}. We emphasize that this is a proxy for identity proliferation rather than a standard identity-switch count---the latter is reported in Table~\ref{tab:mot} for the five ground-truth-annotated sequences. It nevertheless indicates that the wrap-continuous angular state, together with the depth-fused filter, reduces track fragmentation when targets traverse large angular ranges, pass through distortion-prone regions, or experience partial occlusion.

The smaller improvement in \textit{Aerial-Indoor} ($27 \rightarrow 7$) is attributed mainly to intermittent detector failures---the lower body of one participant is occasionally missed, causing abrupt bounding-box variations and temporary track fragmentation---rather than to the tracking model itself.

Overall, the framework improves tracking robustness across both platforms and both environments: by preserving directional continuity through the panorama seam on $\mathbb{S}^2$ and incorporating depth into a unified state-space model, it achieves more stable trajectory estimation and better identity preservation on our self-collected panoramic sequences.

Finally, the formulation provides a unified interface between 2D and 3D tracking: without depth it behaves as a geometry-consistent panoramic 2D tracker, and with LiDAR-derived depth it naturally extends to metric 3D tracking. This flexibility suggests that the spherical state-space model may serve as a general foundation for panoramic perception across heterogeneous sensing platforms, subject to further validation on public benchmarks and larger-scale data.

\section{Conclusion}
We presented a panoramic multi-object tracking framework that integrates a motorized rotating LiDAR with a quad-camera rig, built on a geometry-consistent state representation on the unit sphere: Object direction is parameterized by a two-degree-of-freedom tangent-plane model that avoids the singularities of image-plane tracking and the redundancy of Euclidean 3D representations. An extended spherical Kalman filter jointly estimates direction, scale, aspect ratio, and depth dynamics, providing a unified interface for 2D and 3D tracking.
A controlled ablation shows that the performance gains arise primarily from the wrap-continuity of the angular state and the proposed fusion mechanisms, including LiDAR depth as a joint observation with the modeled $h$-$d$ correlation, angular-height coupling, and blind-zone gain decay, rather than from the parameterization itself.
Our experiments demonstrate robust panoramic tracking under wide fields of view, occlusion, and appearance ambiguity, while maintaining geometric consistency and real-time onboard performance.

Future work will explore tighter coupling with low-level rotating-LiDAR odometry; appearance embedding for long-term identity preservation; larger-scale experiments with more targets, clutter, and moving equipment; systematic evaluation on public panoramic benchmarks; a dedicated validation of wearable localization accuracy under dynamic human-worn conditions; and scaling to multi-robot systems with panoramic sensing.

\appendices
\section{Log/Exp Mapping on \texorpdfstring{$\mathbb{S}^2$}{S2}}
\label{app:logexp}

For completeness, we provide explicit formulas to convert between 
the global unit direction $\mathbf{g}_k \in \mathbb{S}^2$, 
its local 2D coordinates $\mathbf{w}_k \in \mathbb{R}^2$, 
and the reference direction $\bar{\mathbf{g}}_k \in \mathbb{S}^2$ with tangent basis $\mathbf{B}_k=[\mathbf{b}_{1,k},\mathbf{b}_{2,k}]$.

\subsection{From Global to Local (Log Map)}
Given $\bar{\mathbf{g}}_k,\mathbf{g}_k \in \mathbb{S}^2$, define the angular displacement
\begin{equation}
\theta_k = \arccos\!\big(\bar{\mathbf{g}}_k^\top \mathbf{g}_k\big), \quad \theta_k \in [0,\pi].
\end{equation}
Project $\mathbf{g}_k$ onto the tangent plane:
\begin{equation}
\mathbf{p}_k = \big(\mathbf{I}-\bar{\mathbf{g}}_k\bar{\mathbf{g}}_k^\top\big)\mathbf{g}_k, \qquad
\hat{\mathbf{p}}_k = \frac{\mathbf{p}_k}{\|\mathbf{p}_k\|}.
\end{equation}
The tangent displacement is
\begin{equation}
\boldsymbol{\xi}_k = \theta_k \, \hat{\mathbf{p}}_k \in T_{\bar{\mathbf{g}}_k}\mathbb{S}^2,
\end{equation}
and the local 2D coordinates are
\begin{equation}
\mathbf{w}_k =
\begin{bmatrix} w_{1,k} \\ w_{2,k} \end{bmatrix}
= \mathbf{B}_k^\top \boldsymbol{\xi}_k.
\end{equation}

\subsection{From Local to Global (Exp Map)}
Given $\mathbf{w}_k \in \mathbb{R}^2$, let
\begin{equation}
\theta_k = \|\mathbf{w}_k\|, \qquad
\mathbf{u}_k = \frac{\mathbf{B}_k \mathbf{w}_k}{\theta_k}, \quad \text{if } \theta_k > 0.
\end{equation}
The exponential map reconstructs the unit direction:
\begin{equation}
\mathbf{g}_k = \mathrm{Exp}_{\bar{\mathbf{g}}_k}(\mathbf{B}_k\mathbf{w}_k)
= \bar{\mathbf{g}}_k \cos\theta_k + \mathbf{u}_k \sin\theta_k.
\end{equation}
For $\theta_k \approx 0$, a first-order approximation is sufficient:
\begin{equation}
\mathbf{g}_k \approx \bar{\mathbf{g}}_k + \mathbf{B}_k \mathbf{w}_k, \quad
\mathbf{g}_k \leftarrow \frac{\mathbf{g}_k}{\|\mathbf{g}_k\|}.
\end{equation}

These mappings provide a consistent conversion between global directions on the sphere and their local 2D coordinates, which is essential for the spherical Kalman filter formulation described in the main text.

\bibliographystyle{IEEEtran}
\bibliography{refs}

% Generated by IEEEtran.bst, version: 1.14 (2015/08/26)
\begin{thebibliography}{10}
\providecommand{\url}[1]{#1}
\csname url@samestyle\endcsname
\providecommand{\newblock}{\relax}
\providecommand{\bibinfo}[2]{#2}
\providecommand{\BIBentrySTDinterwordspacing}{\spaceskip=0pt\relax}
\providecommand{\BIBentryALTinterwordstretchfactor}{4}
\providecommand{\BIBentryALTinterwordspacing}{\spaceskip=\fontdimen2\font plus
\BIBentryALTinterwordstretchfactor\fontdimen3\font minus
  \fontdimen4\font\relax}
\providecommand{\BIBforeignlanguage}[2]{{%
\expandafter\ifx\csname l@#1\endcsname\relax
\typeout{** WARNING: IEEEtran.bst: No hyphenation pattern has been}%
\typeout{** loaded for the language `#1'. Using the pattern for}%
\typeout{** the default language instead.}%
\else
\language=\csname l@#1\endcsname
\fi
#2}}
\providecommand{\BIBdecl}{\relax}
\BIBdecl

\bibitem{jin2024gs}
R.~Jin, Y.~Gao, Y.~Wang, Y.~Wu, H.~Lu, C.~Xu, and F.~Gao, ``{GS-Planner}: A
  {Gaussian}-splatting-based planning framework for active high-fidelity
  reconstruction,'' in \emph{IEEE/RSJ International Conference on Intelligent
  Robots and Systems (IROS)}, 2024, pp. 11\,202--11\,209.

\bibitem{liao2024mobile}
Y.~Liao, S.~Kang, J.~Li, Y.~Liu, Y.~Liu, Z.~Dong, B.~Yang, and X.~Chen,
  ``{Mobile-Seed}: Joint semantic segmentation and boundary detection for
  mobile robots,'' \emph{IEEE Robotics and Automation Letters}, vol.~9, no.~4,
  pp. 3902--3909, 2024.

\bibitem{Singh_2023_ICCV}
A.~Singh, ``Surround-view vision-based {3D} detection for autonomous driving: A
  survey,'' in \emph{IEEE/CVF International Conference on Computer Vision
  Workshops (ICCVW)}, 2023, pp. 3235--3244.

\bibitem{wen2023panacea}
Y.~Wen, Y.~Zhao, Y.~Liu, F.~Jia, Y.~Wang, C.~Luo, C.~Zhang, T.~Wang, X.~Sun,
  and X.~Zhang, ``{Panacea}: Panoramic and controllable video generation for
  autonomous driving,'' in \emph{IEEE/CVF Conference on Computer Vision and
  Pattern Recognition (CVPR)}, 2024, pp. 6902--6912.

\bibitem{li2025graph}
J.~Li, T.-M. Nguyen, M.~Cao, S.~Yuan, T.-Y. Hung, and L.~Xie, ``Graph
  optimality-aware stochastic {LiDAR} bundle adjustment with progressive
  spatial smoothing,'' \emph{IEEE Transactions on Intelligent Transportation
  Systems}, vol.~26, no.~11, pp. 19\,076--19\,091, 2025.

\bibitem{zhou2021intelligent}
X.~Zhou, X.~Xu, W.~Liang, Z.~Zeng, S.~Shimizu, L.~T. Yang, and Q.~Jin,
  ``Intelligent small object detection for digital twin in smart manufacturing
  with industrial cyber-physical systems,'' \emph{IEEE Transactions on
  Industrial Informatics}, vol.~18, no.~2, pp. 1377--1386, 2022.

\bibitem{wan2019blockchain}
J.~Wan, J.~Li, M.~Imran, D.~Li, and {Fazal-e-Amin}, ``A blockchain-based
  solution for enhancing security and privacy in smart factory,'' \emph{IEEE
  Transactions on Industrial Informatics}, vol.~15, no.~6, pp. 3652--3660,
  2019.

\bibitem{fischer2022ccdt}
T.~Fischer, Y.-H. Yang, S.~Kumar, M.~Sun, and F.~Yu, ``{CC-3DT}: Panoramic {3D}
  object tracking via cross-camera fusion,'' in \emph{Proceedings of the 6th
  Conference on Robot Learning (CoRL)}, ser. PMLR, vol. 205, 2023, pp.
  2294--2305.

\bibitem{yang2026agi2p}
Y.~Yang, J.~Li, Y.~Liao, Y.~Li, R.~Niu, Y.~Zhang, Z.~Dong, B.~Yang, and
  N.~El-Sheimy, ``{AGI2P}: Benchmarking aerial--ground image-to-point cloud
  localization with a large-scale dataset,'' \emph{ISPRS Journal of
  Photogrammetry and Remote Sensing}, vol. 236, pp. 22--36, 2026.

\bibitem{li2026aeos}
J.~Li, X.~Xu, Z.~Liu, S.~Yuan, M.~Cao, and L.~Xie, ``{AEOS}: Active
  environment-aware optimal scanning control for {UAV} {LiDAR}-inertial
  odometry in complex scenes,'' \emph{ISPRS Journal of Photogrammetry and
  Remote Sensing}, vol. 232, pp. 476--491, 2026.

\bibitem{zhang2022bytetrack}
Y.~Zhang, P.~Sun, Y.~Jiang, D.~Yu, F.~Weng, Z.~Yuan, P.~Luo, W.~Liu, and
  X.~Wang, ``{ByteTrack}: Multi-object tracking by associating every detection
  box,'' in \emph{European Conference on Computer Vision (ECCV)}, ser. LNCS,
  vol. 13682, 2022, pp. 1--21.

\bibitem{sun2024smmpod}
J.~Zhang, Y.~Yang, Z.~Gao, H.~Shi, and Y.~Xie, ``{SMM-POD}: Panoramic {3D}
  object detection via spherical multi-stage multi-modal fusion,'' \emph{Remote
  Sensing}, vol.~17, no.~12, 2025, {Art.} no. 2089.

\bibitem{Ong}
J.~Ong, B.-T. Vo, B.-N. Vo, D.~Y. Kim, and S.~Nordholm, ``A {Bayesian} filter
  for multi-view {3D} multi-object tracking with occlusion handling,''
  \emph{IEEE Transactions on Pattern Analysis and Machine Intelligence},
  vol.~44, no.~5, pp. 2246--2263, 2022.

\bibitem{weng2020ab3dmot}
X.~Weng, J.~Wang, D.~Held, and K.~Kitani, ``{3D} multi-object tracking: A
  baseline and new evaluation metrics,'' in \emph{IEEE/RSJ International
  Conference on Intelligent Robots and Systems (IROS)}, 2020, pp.
  10\,359--10\,366.

\bibitem{geyer2000eccv}
C.~Geyer and K.~Daniilidis, ``A unifying theory for central panoramic systems
  and practical implications,'' in \emph{European Conference on Computer Vision
  (ECCV)}, ser. LNCS, vol. 1843, 2000, pp. 445--461.

\bibitem{geyer2001ijcv}
{\reviewed{C.~Geyer and K.~Daniilidis}}, ``Catadioptric projective geometry,''
  \emph{International Journal of Computer Vision}, vol.~45, no.~3, pp.
  223--243, 2001.

\bibitem{scaramuzza2006iros}
D.~Scaramuzza, A.~Martinelli, and R.~Siegwart, ``A toolbox for easily
  calibrating omnidirectional cameras,'' in \emph{IEEE/RSJ International
  Conference on Intelligent Robots and Systems (IROS)}, 2006, pp. 5695--5701.

\bibitem{kiode}
K.~Koide, S.~Oishi, M.~Yokozuka, and A.~Banno, ``General, single-shot,
  target-less, and automatic {LiDAR}-camera extrinsic calibration toolbox,'' in
  \emph{IEEE International Conference on Robotics and Automation (ICRA)}, 2023,
  pp. 11\,301--11\,307.

\bibitem{geiger2012cvpr}
A.~Geiger, P.~Lenz, and R.~Urtasun, ``Are we ready for autonomous driving?
  {The} {KITTI} vision benchmark suite,'' in \emph{IEEE Conference on Computer
  Vision and Pattern Recognition (CVPR)}, 2012, pp. 3354--3361.

\bibitem{caesar2020cvpr}
H.~Caesar, V.~Bankiti, A.~H. Lang, S.~Vora, V.~E. Liong, Q.~Xu, A.~Krishnan,
  Y.~Pan, G.~Baldan, and O.~Beijbom, ``{nuScenes}: A multimodal dataset for
  autonomous driving,'' in \emph{IEEE/CVF Conference on Computer Vision and
  Pattern Recognition (CVPR)}, 2020, pp. 11\,618--11\,628.

\bibitem{li2025limo}
J.~Li, Z.~Liu, X.~Xu, X.~Qin, J.~Liu, S.~Yuan, X.~Fang, and L.~Xie,
  ``{LiMo-Calib}: On-site fast {LiDAR}-motor calibration for quadruped
  robot-based panoramic {3D} sensing system,'' in \emph{IEEE/RSJ International
  Conference on Intelligent Robots and Systems (IROS)}, 2025, pp. 8757--8764.

\bibitem{li2025ua}
J.~Li, X.~Xu, J.~Liu, K.~Cao, S.~Yuan, and L.~Xie, ``{UA-MPC}:
  Uncertainty-aware model predictive control for motorized {LiDAR} odometry,''
  \emph{IEEE Robotics and Automation Letters}, vol.~10, no.~4, pp. 3652--3659,
  2025.

\bibitem{jiao2021mloam}
J.~Jiao, H.~Ye, Y.~Zhu, and M.~Liu, ``Robust odometry and mapping for
  multi-{LiDAR} systems with online extrinsic calibration,'' \emph{IEEE
  Transactions on Robotics}, vol.~38, no.~1, pp. 351--371, 2022.

\bibitem{zhou2020}
X.~Zhou, V.~Koltun, and P.~Kr{\"a}henb{\"u}hl, ``Tracking objects as points,''
  in \emph{European Conference on Computer Vision (ECCV)}, ser. LNCS, vol.
  12349, 2020, pp. 474--490.

\bibitem{sturm2012iros}
J.~Sturm, N.~Engelhard, F.~Endres, W.~Burgard, and D.~Cremers, ``A benchmark
  for the evaluation of {RGB-D} {SLAM} systems,'' in \emph{IEEE/RSJ
  International Conference on Intelligent Robots and Systems (IROS)}, 2012, pp.
  573--580.

\bibitem{burri2016ijrr}
M.~Burri, J.~Nikolic, P.~Gohl, T.~Schneider, J.~Rehder, S.~Omari, M.~W.
  Achtelik, and R.~Siegwart, ``The {EuRoC} micro aerial vehicle datasets,''
  \emph{The International Journal of Robotics Research}, vol.~35, no.~10, pp.
  1157--1163, 2016.

\bibitem{li2023whu}
J.~Li, W.~Wu, B.~Yang, X.~Zou, Y.~Yang, X.~Zhao, and Z.~Dong, ``{WHU-Helmet}: A
  helmet-based multisensor {SLAM} dataset for the evaluation of real-time {3-D}
  mapping in large-scale {GNSS}-denied environments,'' \emph{IEEE Transactions
  on Geoscience and Remote Sensing}, vol.~61, pp. 1--16, 2023.

\bibitem{li2024hcto}
J.~Li, S.~Yuan, M.~Cao, T.-M. Nguyen, K.~Cao, and L.~Xie, ``{HCTO}:
  Optimality-aware {LiDAR} inertial odometry with hybrid continuous time
  optimization for compact wearable mapping system,'' \emph{ISPRS Journal of
  Photogrammetry and Remote Sensing}, vol. 211, pp. 228--243, 2024.

\bibitem{xu2022fast}
W.~Xu, Y.~Cai, D.~He, J.~Lin, and F.~Zhang, ``{FAST-LIO2}: Fast direct
  {LiDAR}-inertial odometry,'' \emph{IEEE Transactions on Robotics}, vol.~38,
  no.~4, pp. 2053--2073, 2022.

\bibitem{luiten2021hota}
J.~Luiten, A.~O{\v{s}}ep, P.~Dendorfer, P.~Torr, A.~Geiger, L.~Leal-Taix{\'e},
  and B.~Leibe, ``{HOTA}: A higher order metric for evaluating multi-object
  tracking,'' \emph{International Journal of Computer Vision}, vol. 129, no.~2,
  pp. 548--578, 2021.

\end{thebibliography}
% R2 layout: match biography photo slots to the existing 0.75 x 0.94 in images.
% Keep all biography text and image sizes unchanged; replace manual negative skips.
\makeatletter
\def\@IEEEBIOphotowidth{0.75in}
\def\@IEEEBIOphotodepth{0.94in}
\def\@IEEEBIOhangwidth{0.84in}
\def\@IEEEBIOhangdepth{0.94in}
\def\@IEEEBIOskipN{0.25\baselineskip}
\makeatother

\begin{IEEEbiography}[{\includegraphics[width=0.75in,height=0.94in,clip,keepaspectratio]{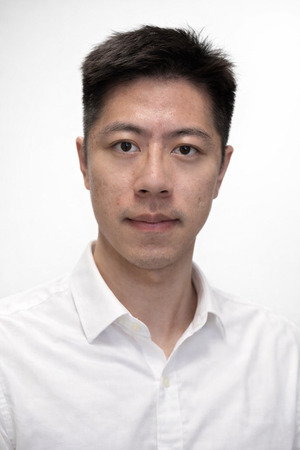}}]
{Zhongyuan Liu}
received the B.Sc. (Hons.) degree in telecommunications engineering with management from Queen Mary University of London, U.K., and Beijing University of Posts and Telecommunications, China, in 2024. He is currently pursuing the M.S. degree in electronics with the School of Electrical and Electronic Engineering, Nanyang Technological University, Singapore.

\end{IEEEbiography}

\begin{IEEEbiography}[{\includegraphics[width=0.75in,height=0.94in,clip,keepaspectratio]{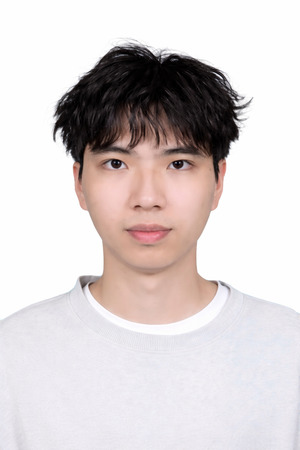}}]
{Shaonan Yu} received the B.Eng. degree in electrical engineering and automation from Nanchang University, Nanchang, China, in 2025. He is currently pursuing the M.S. degree in computer control and automation with the School of Electrical and Electronic Engineering, Nanyang Technological University, Singapore. 
\end{IEEEbiography}

\begin{IEEEbiography}[{\includegraphics[width=0.75in,height=0.94in,clip,keepaspectratio]{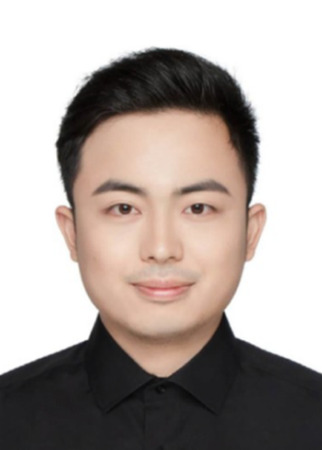}}]
{Jianping Li (Senior Member, IEEE)}
received the B.S. degree in geographic information system (GIS) and the Ph.D. degree in photogrammetry and remote sensing from Wuhan University, Wuhan, China, in 2015 and 2021, respectively. He is currently a Research Fellow with the School of Electrical and Electronic Engineering, Nanyang Technological University, Singapore. His research interests include 3D perception and point cloud data processing, with over 50 publications in venues such as ISPRS, TITS, and TGRS. He received the Science and Technology Progress Award in Surveying and Mapping (2019, 2023) and the National Excellent Doctoral Thesis on LiDAR (2022).
\end{IEEEbiography}

\begin{IEEEbiography}[{\includegraphics[width=0.75in,height=0.94in,clip,keepaspectratio]{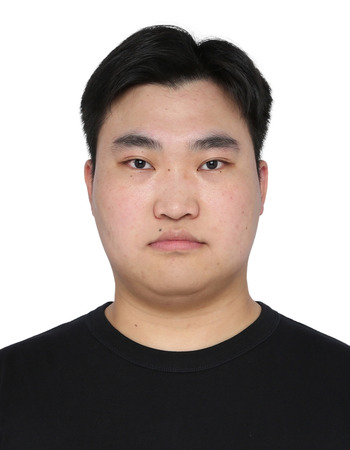}}]
{Pengfei Wan}
received the B.Eng. degree in automation from Northwestern Polytechnical University, Xi’an, China, in 2022. He is currently pursuing the M.S. degree in computer control and automation with the School of Electrical and Electronic Engineering, Nanyang Technological University, Singapore. 
\end{IEEEbiography}

\begin{IEEEbiography}[{\includegraphics[width=0.75in,height=0.94in,clip,keepaspectratio]{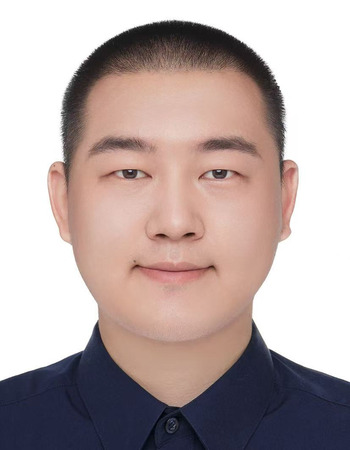}}]
{Xinhang Xu} 
received the B.Eng. degree in automation from the South China University of Technology, Guangzhou, China, and the M.Sc. degree in electrical and electronic engineering from Nanyang Technological University, Singapore, in 2023. He is currently pursuing the Ph.D. degree in electrical and electronic engineering at Nanyang Technological University, Singapore. 
\end{IEEEbiography}

\begin{IEEEbiography}[{\includegraphics[width=0.75in,height=0.94in,clip,keepaspectratio]{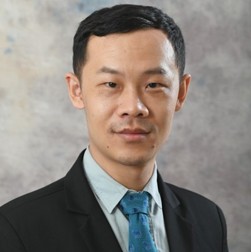}}]
{Pengfei Wang}
received the B.Eng. degree in mechanical engineering from Huazhong University of Science and Technology, China, in 2012, and the Ph.D. degree from the National University of Singapore, Singapore, in 2016.

He is currently a Scientist with ST Engineering, Singapore. His research interests include multi-object tracking, robotic perception, and sensor fusion.
\end{IEEEbiography}

\begin{IEEEbiography}[{\includegraphics[width=0.75in,height=0.94in,clip,keepaspectratio]{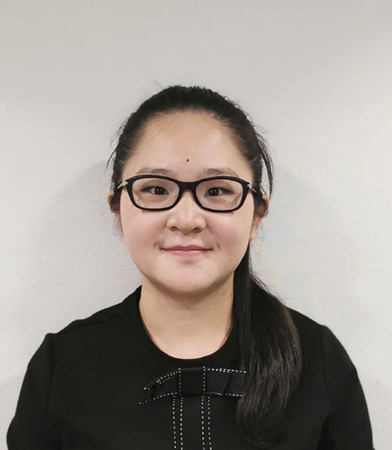}}]
{Maggie Y. Gao (Member, ICE)}
received the B.Mgt. degree in construction management from Central South University, China, in 2012, the M.Sc. degree in civil infrastructural engineering and management from the Hong Kong University of Science and Technology, Hong Kong, in 2013, and the Ph.D. degree in civil engineering from Nanyang Technological University (NTU), Singapore.
She has over nine years of industry experience on major infrastructure and MRT projects in Singapore and Hong Kong, and is currently a Research Fellow with the School of Civil and Environmental Engineering, NTU.
\end{IEEEbiography}

\begin{IEEEbiography}[{\includegraphics[width=0.75in,height=0.94in,clip,keepaspectratio]{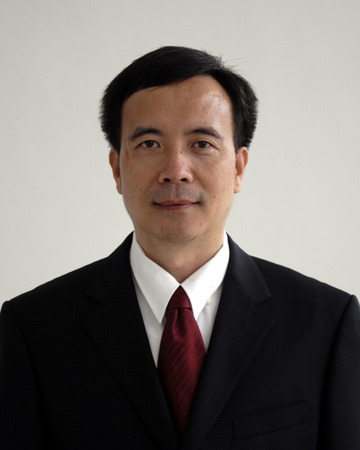}}]
{Lihua Xie (Fellow, IEEE)}
received the Ph.D. degree in electrical engineering from the University of Newcastle, Callaghan, NSW, Australia, in 1992.

He is currently a President's Chair Professor and the Director of the Center for Advanced Robotics Technology Innovation, Nanyang Technological University, Singapore. His research interests include robust control and estimation, networked control systems, multiagent systems, and uncrewed systems.

Dr. Xie is a Fellow of the Academy of Engineering Singapore, IFAC, and CAA. He is currently the Editor-in-Chief of \textit{Unmanned Systems} and has served as an Associate Editor for several journals including \textit{IEEE Transactions on Automatic Control}, \textit{Automatica}, and \textit{IEEE Transactions on Control Systems Technology}.
\end{IEEEbiography}

\clearpage
\onecolumn

% \appendices
\section{Key Parameters and Notation}
\label{app:params}

Table~\ref{tab:notation} summarizes the notation used throughout the paper, and Table~\ref{tab:params} lists the key parameters of the implementation. All parameter values are held fixed across all sequences, platforms, and ablation variants, with no per-sequence tuning.

\begin{figure}[h]
\centering
\includegraphics[width=\linewidth]{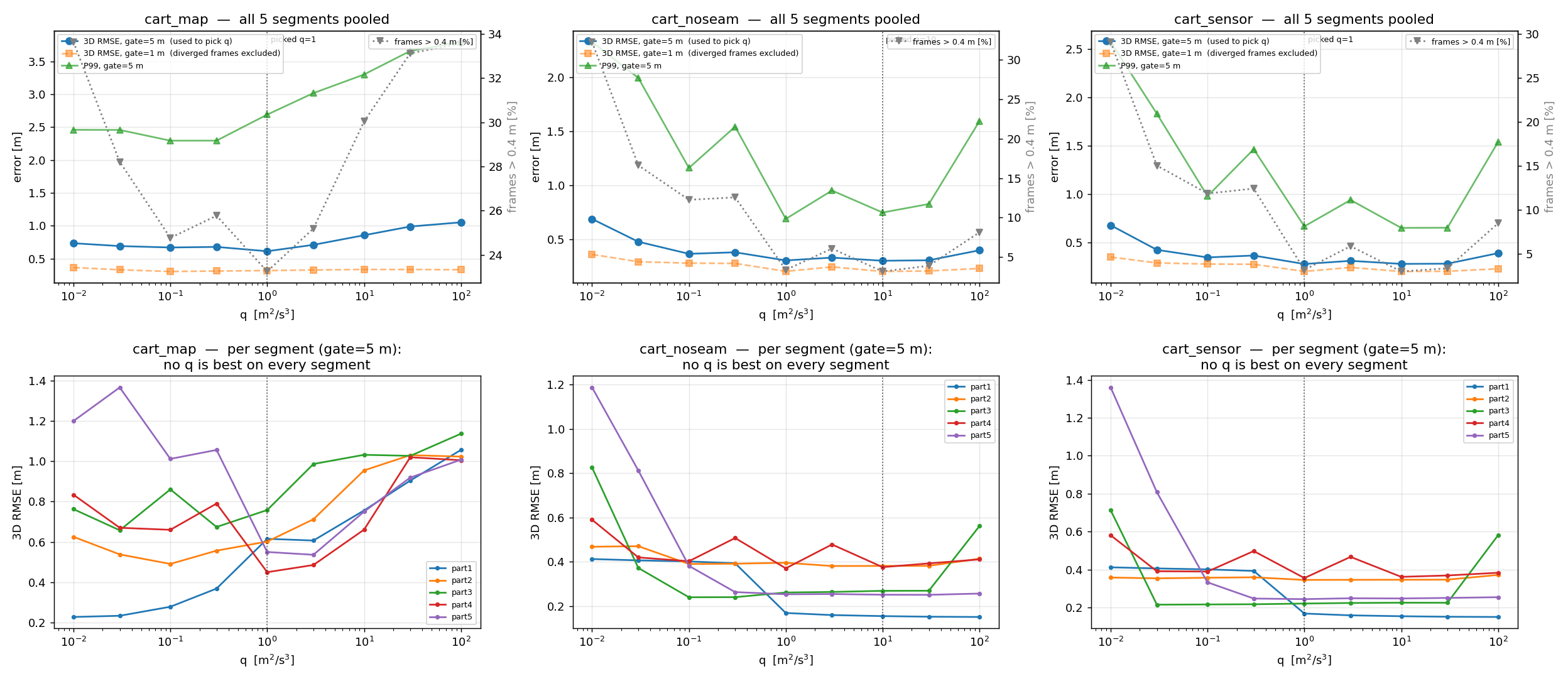}
\caption{Process-noise sweep of the Cartesian baselines over $q\in[0.01,100]$ (five sequences merged and per-sequence): merged 3D RMSE, P99, and the fraction of frames above 0.4\,m, with the selected operating point marked. No single $q$ is optimal across all sequences.}
\label{fig:q_sweep}
\end{figure}

\begin{table}[h]
\centering
\caption{Notation used in the spherical tracking formulation.}
\label{tab:notation}
\footnotesize
\setlength{\tabcolsep}{3.5pt}
\renewcommand{\arraystretch}{1.08}

\begin{tabularx}{\linewidth}{@{}>{\raggedright\arraybackslash}p{0.34\linewidth}X@{}}
\toprule
Symbol & Description \\
\midrule
$\mathbb{S}^2$ & Unit sphere used to represent object bearing directions. \\
$\mathbf{g}_k \in \mathbb{S}^2$ & Object bearing vector at time step $k$. \\
$\bar{\mathbf{g}}_k \in \mathbb{S}^2$ & Reference bearing used to define the local tangent space. \\
$T_{\bar{\mathbf{g}}_k}\mathbb{S}^2$ & Tangent space of $\mathbb{S}^2$ at $\bar{\mathbf{g}}_k$. \\
$\mathbf{B}_k=[\mathbf{b}_{1,k},\mathbf{b}_{2,k}]$ & Orthonormal basis spanning $T_{\bar{\mathbf{g}}_k}\mathbb{S}^2$. \\
$\mathbf{w}_k=[w_{1,k},w_{2,k}]^\top$ & Local tangent-plane coordinates of the bearing state. \\
$\mathbf{x}_k$ & Full target state vector. \\
$\hat{\mathbf{x}}_{k|k-1},\hat{\mathbf{x}}_{k|k}$ & Predicted and corrected state estimates. \\
$\mathbf{P}_{k|k-1},\mathbf{P}_{k|k}$ & Covariances of the predicted and corrected state estimates. \\
$\mathbf{F},\mathbf{Q}$ & State transition matrix and process noise covariance. \\
$\mathbf{z}_k,\mathbf{H},\mathbf{R}$ & Measurement vector, observation matrix, and measurement noise covariance. \\
\bottomrule
\end{tabularx}
\end{table}

The height and distance channels use the standard continuous-white-noise-acceleration (CWNA) process-noise blocks $q\,[\Delta t^{3}/3,\ \Delta t^{2}/2;\ \Delta t^{2}/2,\ \Delta t]$ (units m$^{2}$/s$^{3}$ for $q$), while the direction and angular-aspect-ratio channels use a diagonal random-walk form $Q_{ii}=\sigma_{i}^{2}\,\Delta t$ with $\sigma_{i}$ proportional to the current angular height $\alpha$; the filter step takes the nominal value $\Delta t=1/15$\,s. Figure~\ref{fig:q_sweep} shows the process-noise sweep of the Cartesian baselines over $q\in[0.01,100]$: the merged optimum is $q=1.0$, and no single $q$ is optimal across all five sequences.

% Full-width table: avoids the severe wrapping seen in the current PDF.
\begin{table}[]
\centering
\caption{Key parameters of the implementation.}
\label{tab:params}
\footnotesize
\setlength{\tabcolsep}{6pt}
\renewcommand{\arraystretch}{1.10}

\begin{tabularx}{\textwidth}{@{}>{\raggedright\arraybackslash}p{0.27\textwidth}X@{}}
\toprule
Parameter & Value \\
\midrule
Detector & YOLO11-m, TensorRT FP16, $640\times640$, COCO-pretrained (no fine-tuning) \\
Confidence / NMS threshold & 0.50 / 0.65 \\
Panorama & $1280\times640$ equirectangular, vertical FoV $\pm90^{\circ}$ \\
Camera model & Kalibr omni (Mei) + radtan, 4 fisheye \\
Sensor rates & image 15.1\,Hz; LiDAR 10\,Hz; IMU $\approx$200\,Hz; encoder 167\,Hz \\
\midrule
track / high / match thresholds & 0.30 / 0.50 / 0.80 (low-score 0.50; unconfirmed 0.70) \\
Max track loss & 15 frames ($\approx$1.0\,s) \\
Association cost & $1-$seam-aware image-plane IoU \\
\midrule
Direction proc.\ noise & $\sigma_w=0.0601\,\alpha$, $\sigma_{\dot{w}}=0.0180\,\alpha$ (rad; random-walk form $Q_{ii}=\sigma_i^2\Delta t$) \\
Aspect-ratio proc.\ noise & $\sigma_{\ell}=10^{-2}$, $\sigma_{\dot{\ell}}=10^{-5}$ (random-walk form) \\
Height / distance proc.\ noise & CWNA blocks $q[\Delta t^{3}/3,\ \Delta t^{2}/2;\ \Delta t^{2}/2,\ \Delta t]$, $q_h=10^{-3}$, $q_D=1.0$ (m$^2$/s$^3$) \\
Filter step $T$ & $1/15$\,s (nominal) \\
Direction meas.\ noise & $\sigma_{\mathrm{dir}}=0.0245\,\alpha$ (rad) \\
Angular-aspect-ratio meas.\ noise & $R_{\ell}=10^{-2}$ \\
Angular-height meas.\ noise & $\sigma_{\alpha}=0.0147$\,rad ($R_{\alpha}=2.17\times10^{-4}$\,rad$^2$) \\
Depth meas.\ noise & $R_d=\min(0.03^2+s_{\mathrm{med}}^2+0.15^2,\,0.25)$\,m$^2$ \\
NIS gates & 8.0 (scalar); 15.0 (5-D joint); joint rejection falls back to image-only \\
Blind-zone gain decay & $\tau=1.0$\,s, $w=\exp(-t_{\mathrm{blind}}/\tau)$; $\dot{h}$ frozen for $t_{\mathrm{blind}}>\tau$ \\
Initialization & $\sigma_D=0.25\,d_0$, $\sigma_{\dot{D}}=3$\,m/s, $\sigma_{\dot{h}}=0.5$\,m/s; nominal scale prior 1.73\,m (first frame w/o depth only) \\
Numerical protection & $\mathbf{P}$ diagonal floor $10^{-4}$ \\
\midrule
Depth cluster & $0.3$\,m window, $\ge 125$ splat points, $\ge 15\%$ valid pixels ($5\times5$ splat) \\
\midrule
Sync tolerance / max wait & 60 / 150\,ms \\
Cloud--image content offset & 0.15\,s (calibrated per session) \\
LiDAR--panorama optical-center $z$ offset & 0.25\,m \\
Rotary platform speed & $\approx$6\,RPM (period 9.99\,s) \\
\bottomrule
\end{tabularx}
\end{table}

Table~\ref{tab:mot_full} reports the complete MOT metrics, including FP, FN, fragmentation, and MT/PT/ML, which complement the summary columns of Table~\ref{tab:mot}.

% Full-width table, but no resizebox: preserves readable IEEE-sized text.
\begin{table}[]
\centering
\caption{Complete MOT metrics per sequence and merged ($d_{\max}=1.0$\,m). MOTP is in meters; FP/FN: false-positive/false-negative frames; Frag: fragmentations; MT/PT/ML: numbers of mostly-tracked, partially-tracked, and mostly-lost targets. HOTA is reported together with its detection (DetA) and association (AssA) components, following~\cite{luiten2021hota}.}
\label{tab:mot_full}
\footnotesize
\setlength{\tabcolsep}{3.8pt}
\renewcommand{\arraystretch}{1.08}

\begin{tabular*}{\textwidth}{@{\extracolsep{\fill}}llccccccccccc@{}}
\toprule
Seq. & Method & MOTA & MOTP (m) & IDF1 & HOTA & DetA & AssA & FP & FN & IDS & Frag & MT/PT/ML \\
\midrule
\multirow{2}{*}{Seq1} & Ours & 0.997 & 0.114 & 0.999 & 0.915 & 0.869 & 1.000 & 4 & 0 & 0 & 0 & 3/0/0 \\
 & Baseline & 0.952 & 0.087 & 0.976 & 0.922 & 0.881 & 1.000 & 2 & 73 & 0 & 0 & 3/0/0 \\
\multirow{2}{*}{Seq2} & Ours & 0.989 & 0.173 & 0.994 & 0.854 & 0.810 & 1.000 & 4 & 13 & 0 & 9 & 3/0/0 \\
 & Baseline & 0.907 & 0.260 & 0.540 & 0.520 & 0.623 & 0.468 & 29 & 105 & 5 & 17 & 3/0/0 \\
\multirow{2}{*}{Seq3} & Ours & 0.997 & 0.190 & 0.998 & 0.844 & 0.799 & 1.000 & 3 & 2 & 0 & 2 & 3/0/0 \\
 & Baseline & 0.710 & 0.311 & 0.439 & 0.420 & 0.457 & 0.482 & 120 & 338 & 8 & 14 & 2/1/0 \\
\multirow{2}{*}{Seq4} & Ours & 0.944 & 0.199 & 0.972 & 0.817 & 0.740 & 1.000 & 48 & 34 & 0 & 9 & 3/0/0 \\
 & Baseline & 0.898 & 0.224 & 0.498 & 0.530 & 0.681 & 0.476 & 53 & 91 & 6 & 14 & 3/0/0 \\
\multirow{2}{*}{Seq5} & Ours & 0.991 & 0.211 & 0.995 & 0.822 & 0.758 & 1.000 & 5 & 2 & 0 & 1 & 3/0/0 \\
 & Baseline & 0.827 & 0.313 & 0.904 & 0.671 & 0.517 & 0.972 & 55 & 76 & 1 & 8 & 3/0/0 \\
\midrule
\multirow{2}{*}{All} & Ours & \textbf{0.983} & \textbf{0.173} & \textbf{0.992} & \textbf{0.856} & 0.794 & \textbf{1.000} & 64 & 51 & \textbf{0} & 21 & 15/0/0 \\
 & Baseline & 0.860 & 0.227 & 0.649 & 0.623 & 0.627 & 0.697 & 259 & 683 & 20 & 53 & 14/1/0 \\
\bottomrule
\end{tabular*}
\end{table}
\end{document}